\documentclass[final,3p,times,authoryear]{elsarticle}

\usepackage{booktabs}
\usepackage{threeparttable}

\usepackage{natbib}
\usepackage{rotating}

\usepackage{caption} 

\usepackage{amsthm}
\usepackage{amsmath}
\usepackage{amssymb}
\usepackage{bm}

\usepackage{dsfont}
\usepackage{graphicx}
\usepackage{multirow}
\usepackage{subcaption}
\usepackage{hyperref} 

\usepackage{algorithm}
\usepackage{algpseudocode}

\usepackage{circledsteps}

\newtheorem{definition}{Definition}[section] 
\newtheorem{assumption}{Assumption}[section]
\begin{document}
	
	\begin{frontmatter}
		
		
		
		\title{Self Balancing Neural Network: A Novel Method to Estimate Average Treatment Effect} 
		
		
		\author[label1,label2]{Atomsa Gemechu Abdisa} 
		\ead{gemechu.abdisa@aau.edu.et}
		\author{Yingchun Zhou\corref{cor1}\fnref{label1}}
		\ead{yczhou@stat.ecnu.edu.cn}
		\cortext[cor1]{Corresponding author.}
		\author[label1]{Yuqi Qiu}
		\ead{yqqiu@fem.ecnu.edu.cn}
		\affiliation[label1]{organization={KLATASDS-MOE, School of Statistics, East China Normal University},
			state={Shanghai},
			country={People’s Republic of China}}
		\affiliation[label2]{organization={Department of Statistics, College of Natural and Computational Sciences, Addis Ababa University},
			city={Addis Ababa},
			postcode={1176}, 
			country={Ethiopia}}
		\begin{abstract}
			In observational studies, confounding variables affect both treatment and outcome. Moreover, instrumental variables also influence the treatment assignment mechanism. This situation sets the study apart from a standard randomized controlled trial, where the treatment assignment is random. Due to this situation, the estimated average treatment effect becomes biased. To address this issue, a standard approach is to incorporate the estimated propensity score when estimating the average treatment effect. However, these methods incur the risk of misspecification in propensity score models. To solve this issue, a novel method called the "Self balancing neural network" (Sbnet), which lets the model itself obtain its pseudo propensity score from the balancing net, is proposed in this study. The proposed method estimates the average treatment effect by using the balancing net as a key part of the feedforward neural network. This formulation resolves the estimation of the average treatment effect in one step. Moreover, the multi-pseudo propensity score framework, which is estimated from the diversified balancing net and used for the estimation of the average treatment effect, is presented. Finally, the proposed methods are compared with state-of-the-art methods on three simulation setups and real-world datasets. It has been shown that the proposed self-balancing neural network shows better performance than state-of-the-art methods.
		\end{abstract}

		\begin{keyword}
			Average treatment effect, Self balancing, Neural network, Pseudo propensity score 
		\end{keyword}
		
	\end{frontmatter}
	
		

		\section{Introduction}
		\label{sec1}
		Observational studies have become a primary focus within the field of causal inference. This study takes place when randomization is impractical due to reasons such as ethical considerations, cost, and other constraints. But still, randomized controlled trials (RCTs) are a gold standard for this study \citep{rosenbaum1983central}. Therefore, it finds widespread application in fields like clinical and marketing research.
		
		Unlike in randomized experiments, data collected from observational studies usually face a confounding problem, where some or all of the variables that affect the outcome also have an effect on treatment. In addition to this, instrumental variables also play a role in the mechanism of treatment assignment. This situation disrupts the independent distribution of treatment assignments. As a result, the estimation of the average treatment effect becomes biased if one constructs the model for this study design in a similar way as one can do under a randomized controlled trial.

		To address this problem, often a propensity score is estimated and used for propensity score matching (PSM), inverse probability weighting (IPW), or stratification \citep{rosenbaum1983central, imbens2015causal,pearl1993bayesian,pearlbook}. The propensity score can be estimated using common classification methods, such as logistic regression or machine learning models \citep{rosenbaum1983central,guo2020propensity}.  The literature has also explored machine learning models to capture non-linearity structure in the dataset. For instance, researchers introduced DragonNet as an extension of TARNet, utilizing propensity score concepts for causal estimation \citep{shi2019adapting, shalit2017estimating}. 
		
		In estimating propensity scores, it's important to consider confounding variables, predictors, and strong instrumental variables \citep{brookhart2006variable,patrick2011implications,schuster2016propensity,rubin1996matching,myers2011effects,lunceford2004stratification,rubin1997estimating}. Identifying these variables from pre-treatment variables is one of the key challenges in modelling a propensity score model. Moreover, obtaining a correct model specification for propensity scores is often a complex task. Hence, propensity score models are often vulnerable to model misspecification, which can introduce biases into causal estimates.
		
		To solve this problem, doubly robust methods have been suggested, assuming that at least one of the two models—the propensity score model or the outcome model—is set up correctly \citep{bang2005doubly}. This approach ensures more reliable causal inference, even when one of the models is misspecified. Furthermore, the doubly robust principle has been integrated into machine learning, leading to the development of doubly robust learning methods such as double machine learning (DML) \citep{chernozhukov2018double, knaus2022double}. Although these methods require that one of the models be correctly specified, establishing this requirement is not simple. Moreover, double machine learning models are subject to plug-in bias. \citep{Hernan2024-HERCIW,JMLR:v25:22-1233}.
		
		To address bias due to model misspecifications for the propensity score model, this paper introduces a one-step solution to handle propensity scores and outcome models jointly. We refer to this method as the "Self-Balancing Neural Network" (Sbnet). The design of the Sbnet model allows it to update its own balancing net within itself, thereby reducing the final loss. In this design, the integrated architecture of the balancing net forms a pseudo propensity score.
		
		In Sbnet, the outcome net integrates the balancing net, which eliminates the need for separate model identification and estimation. The outcome net incorporates the balancing net as one of its inputs and shares gradients with the input layer during back-propagation to optimize its parameters. This design differs from current methods that utilize propensity score modeling because it does not use the pseudo propensity score, which is obtained from its balancing net, to create treatment loss nor to penalize the objective function of the outcome model. Thus, Sbnet creates a dynamic regression adjustment model by utilizing the characteristics of its balancing nets. This method makes it easier to address misspecification in the propensity score model because the pseudo propensity score is created by the balancing net, which updates during the training process of Sbnet.	
		
		\textbf{{The main contributions of this paper}} are: \Circled{1} It presents a self-balancing neural network that estimates the average treatment effect by using the pseudo propensity score as a key part of the feedforward neural network. The Sbnet has nested the balancing net inside its architecture to obtain the pseudo propensity score. This feature makes the proposed method a one-step solution for handling the estimation of both the pseudo propensity score and the average treatment effect simultaneously. \Circled{2} The paper also presents the concept of multiple pseudo propensity scores. In this case, multiple pseudo propensity scores are estimated from the diversified balancing net and used for the estimation of the average treatment effect. Similar to other ensemble learning techniques, this method aims to improve the estimation of the model. 
		
		The structure of this paper is as follows: Section \ref{sec1} is the introduction section for the self-balancing neural network. In Section \ref{sec2}, notations, definitions, and assumptions important to the paper are provided. In Section \ref{sec3}, the self-balancing neural network method is provided in detail. A simulation study and application of the proposed method are provided in Sections \ref{sec4} and \ref{sec5}, respectively. Finally, Section \ref{sec6} provides the conclusion section.
		
		\section{Notations, Definitions and Assumptions}
		\label{sec2}
		Let the triple \(\{\mathbb{X},T,Y\}\) denote the set of pre-treatment variables, treatment, and outcome from the space  \(\{\mathcal{X},\mathcal{T},\mathcal{Y}\}\). Moreover, let $n$ denote the sample size and $k$ denote the number of pre-treatment variables. The values of $\mathcal{X}$ are defined on $\mathbb{R}^{n\times k}$; the binary space $\mathcal{T}$ is defined as $\mathcal{T}\in \{0,1\}^n$, and $\mathcal{Y}$ takes values on $\mathcal{R}^n$. 
		
		Note that in the triple, only pre-treatment variables are considered. Often, collider effects arise when one consider post-treatment variables. This occurs because both the treatment and the outcome can influence the post-treatment variable. This circumstance is usually known as Berkson's paradox \citep{berkson1946limitations}. Inclusion of such variables results in a biased estimated value. This paper does not address the potential collider effect problem. Hence, $\mathbb{X}$ is chosen as a set of pre-treatment variables.  
		
		The subset of pre-treatment variables ($\mathbb{X}$) may affect treatment assignment or outcome variables. Those variables that affect both outcome and treatment are known as confounders. Those that affect only treatment assignment are known as instrumental variables. Moreover, those that affect only outcome variables are called predictors, and the rest of the variables, which have no effect on both treatment and outcome, are called spurious variables.
		
		Using the values obtained from the triple, an outcome regression model is proposed to estimate the average treatment effect on the outcome. In general, this model is defined as follows:
		\begin{equation}
			Y(T)=f_{g}(T,\mathbb{X}) =f(T,\mathbb{X},g(\mathbb{X}))\label{outconereg}
		\end{equation}
		where $f(\cdot)$ denotes functional form of outcome regression model. It is common to use models such as linear regression and machine learning models as the outcome regression model. Moreover $g(\mathbb{X})$ denotes the propensity score, which is usually estimated based on propensity score models.
		\begin{definition}
			Propensity Score is the probability that treatment $T=1$ is assigned to units according to some pre-treatment features $\mathbb{X}$.
		\end{definition}
		
		To estimate the propensity score, models such as logistic regression and machine learning models for classification are used. However, as stated earlier, it is not an easy task to correctly specify the propensity score model and identify variables to include in the model. \citet{pearlbook} also has questions about how to justify which subsets of pre-treatment variables result in bias reduction when they are used in a propensity score model.
		
		This paper introduces the pseudo propensity score approach to resolve problems, such as model misspecification, associated with the propensity score model. In contrast to estimating the propensity score, the Sbnet generates the pseudo propensity score, making it a crucial component of the outcome model. The proposed method shares the same outcome regression model $f(T,\mathbb{X},g(\mathbb{X}))$ given in Equation \ref{outconereg}. 
		
		In the Sbnet architecture, $g(\mathbb{X})$ is its balancing net, which is nested inside the outcome model.  This component helps to estimate the pseudo propensity score, replacing the role of the propensity score model. Since $\mathbb{X}$ is a balancing score, the resulting pseudo propensity score from $g(\mathbb{X})$ becomes a balancing score for the outcome model in Sbnet architecture. This follows the fact that based on a balancing score one can obtain other balancing score \citep{rosenbaum1983central}. The pseudo propensity score is obtained dynamically from the nested architecture inside Sbnet, following certain data-generating mechanisms.

		The estimation of average treatment effect using self-balancing neural networks requires similar assumptions in any other outcome regression model. These assumptions are expressed below in Assumptions \ref{assmp1} to \ref{assmp4}.
		
		\begin{assumption}\label{assmp1}
			There is no unmeasured confounders.
		\end{assumption}
		
		Confounders are crucial variables derived from the pre-treatment data. When there are hidden (or missing) confounders, the resulting estimation will be biased. Hence, like other causal inference models, Sbnet requires that all confounding variables are not hidden. 
		
		Like the propensity score, the pseudo propensity score is a balancing score which is a function of pre-treatment variables. Hence we assume that conditional independence between treatment assignment and outcome variable holds given the balancing score of Sbnet, which is a pseudo propensity score.
		
		\begin{assumption}\label{assmp2}  Given a balancing score $g(\mathbb{X})$, outcome variable is independent of treatment assignment:
			$$Y(t) \perp T | g(\mathbb{X}), \quad \text{ where } t=0,1.$$
		\end{assumption}

		When a balancing score function ($g(\mathbb{X})$) is properly created, the way we measure outcomes becomes independent of the treatment assignment mechanism when we consider $g(\mathbb{X})$. This results in conditional exchangeability between the treated and the untreated. Under this assumption, we are simply appealing to de Finetti's theorem of the row exchangeability condition \citep{deFinetti1979-DEFTOP-2}. Properly estimating the propensity score is an important task to make sure this assumption holds. This requires appropriately identifying these subsets and model specification to model the propensity score. When these are satisfied and propensity is correctly estimated, \citet{rosenbaum1983central} emphasizes that this assumption will hold.
		
		\begin{assumption}\label{assmp3}
			Positivity: Any unit in the study has a nonzero probability $(e(\mathbb{X}_i))$ to receive the treatment $T=1$. i.e., $0<e(\mathbb{X}_i)<1 ~ \forall_{i\in[1,n]}$.
		\end{assumption}
		
		For binary treatment, $t=0,1$, they can be assigned to any units (or individuals) in the study with non zero probability. Under this assumption, treatment effects be can computed from outcome variable, which is defined as follows:
		\begin{equation}
			Y(t)=\begin{cases}
				Y(0) & \text{if } t=0\\
				Y(1) & \text{if } t=1
			\end{cases} \label{outcome1}
		\end{equation}
		where $Y(0)$ and $Y(1)$ are potential outcomes. 
		
		Conditions stated in Assumptions \ref{assmp2} and \ref{assmp3} are usually referred to as the strong ignorability condition. To establish these conditions, Assumption \ref{assmp1} is the key step. Under these assumptions, one can estimate treatment effects by comparing the outcomes of binary treatments. In this paper, the main interest is to compute the average treatment effect (ATE). For the outcome model in Equation \ref{outconereg}, the average treatment effect ($\gamma$) can be computed as follows:
		
		\begin{equation}
			\gamma= \mathbb{E}\left(f_{g}(T=\boldsymbol{1},\mathbb{X})\right)-\mathbb{E}\left(f_{g}(T=\boldsymbol{0},\mathbb{X})\right). \label{ate1}
		\end{equation}
		Its estimated value is computed  as follows
		\begin{equation}
			\hat{\gamma}= \frac{1}{n}\sum_{i=1}^{n}\left(\hat{f}_{\hat{g}}(T_i=1,\mathbb{X}_i)\right)-\frac{1}{n}\sum_{i=1}^{n}\left(\hat{f}_{\hat{g}}(T_i=0,\mathbb{X}_i)\right). \label{ate2}
		\end{equation}
		
		The estimation of ATE is reasonable when the treatment effect is homogenous. Hence we assume that treatment effect is homogeneous. 
		
		\begin{assumption}\label{assmp4}
			Treatment effect is homogeneous.
		\end{assumption}

		These are basic assumptions needed to form a model that can estimate the average treatment effect. However, some of these assumptions cannot be tested directly \citep{imbens2015causal}. In general, we assume that these assumptions hold so that we can form a valid model and interpret the average treatment effect.
		
		In the following section, we propose a novel neural network model that incorporates a pseudo propensity score to estimate the average treatment effect based on these assumptions. Novel concepts are also presented in the section to establish how to guarantee the convergence of pseudo propensity score generation.
		
		\section{Self Balancing Neural Network}
		\label{sec3}
		Self-Balancing Neural Network (Sbnet) is designed as a one-step solution to the functional relationship defined in Equation \ref{outconereg}. It uses its own pseudo propensity score to balance itself rather than relying on a separate model for the propensity score, which often suffers from misspecification. During the optimisation process, the values of the pseudo propensity score and the outcome will be jointly estimated, holding that the pseudo propensity score is one of the predictors for the outcome, alongside pre-treatment variables and treatment. 
		
		The following subsections present how a standard feedforward neural network is modified to design Sbnet and then backpropagated to update its weights and bias. To make the structure of Sbnet understood, there is no penalty term included in loss functions to regularise the architecture. A method to assess convergence for the estimation process of pseudo propensity scores is also presented.

		\subsection{General Architecture of Sbnet}

		A self-balancing neural network is proposed by enhancing a fully connected feedforward neural network of the outcome model to have a balancing net, which generates (estimates) a pseudo-propensity score. This balancing net is nested inside the architecture and aims to balance the outcome net. Both forward propagation and backpropagation are done for both nets as one standard multilayer perceptron.
		
		Unlike the supervised learning methods, the balancing net has no target to train on. Rather, it works as an autonomous agent, from which a pseudo propensity score is obtained, to improve estimation of the average treatment effect using Sbnet. This concept is discussed in Subsection \ref{optimizationstage} when optimization of Sbnet is discussed.

		\begin{figure}[h!]
			\centering
			\includegraphics[width=0.5\textwidth]{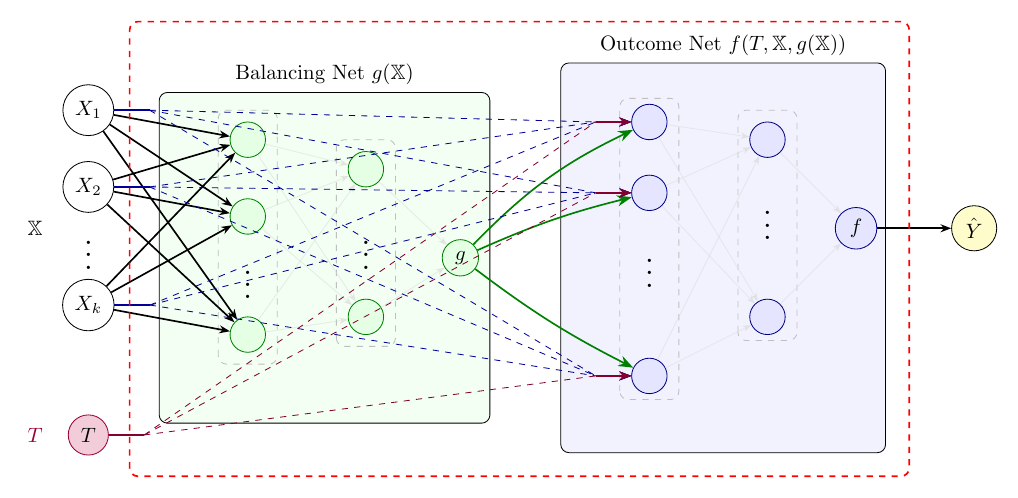}
			\caption{Self Balancing Neural Network architecture. Both arrows with broken lines and without broken lines indicate the forward propagation from input to hidden layer. The broken lines are used only for the sake of visibility.}
			\label{Sbnet}
		\end{figure}
		
		In Figure \ref{Sbnet}, the general feedforward architecture of the self-balancing neural network is given. First, the pre-treatment variable $\mathbb{X}$ feeds forward to the balancing net, then the retained pseudo propensity from the outer layer of the balancing net, combined with the treatment $T$ and the pre-treatment variable $\mathbb{X}$, is fed forward to the outcome net. In this way, the forward propagation step will be done inside the balancing and outcome nets of the Sbnet architecture. Like in standard neural network models, optimal architecture for Sbnet can be easily obtained based on grid search. One can also use one hidden layer for both balancing and outcome nets, then search for optimal hyperparameters by grid searching.
		
		To further highlight the feedforward process, let us start with how it will be done inside the balancing net. During the feedforward process, the pre-treatment variables feed into the first hidden layer, and then the result from the first hidden layer maps to the next layer. Each layer has a pre-defined number of nodes/neurones. Assume the first hidden layer has $l$ number of neurones/nodes; then $\mathbb{X}$ maps to/feeds forward to each of the neurones. This step can be described as follows:
		\begin{equation}
			\resizebox{0.9\textwidth}{!}{$
				\underbrace{\left(X_1,X_2,\cdots,X_k\right)}_{=\mathbb{X}} \to \underbrace{\left\{g_1\left(\sum_{j=1}^{k}\beta_{j}^{(1,1)}X_j+b^{(1,1)}\right), g_1\left(\sum_{j=1}^{k}\beta_{j}^{(2,1)}X_j+b^{(2,1)}\right),\cdots, g_1\left(\sum_{j=1}^{k}\beta_{j}^{(l,1)}X_j+b^{(l,1)}\right)\right\}}_{\text{for } l \text{ nodes}}$
			}\label{thefirstfeedforward}
		\end{equation}
		where $g_1$ represents the activation function for the first hidden layer, $\left\{\beta_{j}^{(s,1)}\right\}_{j=1}^k$ and $b^{(s,1)}$ denote the weights and biases used to map the input to the $s$ node of the first hidden layer. Using the activation function, the resulting weighted combination of variables in $\mathbb{X}$ plus the bias component will be transferred. This activation function works similarly to link functions used in generalized linear models. The result in \ref{thefirstfeedforward} can be further simplified and written as follows:
		
		\begin{equation}
			\mathbb{X} \to \left\{g_1\left(\sum_{j=1}^{k}\beta_{j}^{(s,1)}X_j+b^{(s,1)}\right)\right\}_{s=1}^{l}.
		\end{equation}
		
		Let $\mathbf{u}_s^{(1)}=\left(\beta_{1}^{(s,1)},\beta_{2}^{(s,1)},\cdots, \beta_{k}^{(s,1)}\right)^{\top}$, $\mathbf{U}_g^{(1)}=\left\{\mathbf{u}^{(s,1)}\right\}_{s=1}^l$ and $\mathbf{b}^{(1)}_g=\left\{b^{(s,1)}\right\}_{s=1}^l$, and denote the input pre-treatment variable by $\mathbf{a}_g^{(0)}$, i.e.  $\mathbf{a}_g^{(0)}=\mathbb{X}$. Moreover, if $\mathbf{a}_g^{(1)}$ denotes the resulting vectors from the first hidden layer, then the forward propagation from input to the first hidden layer will be put together as follows: 
		\begin{equation}
			\mathbf{a}_g^{(0)}\to \mathbf{a}_g^{(1)} = \left\{g_1(\mathbf{u}_s^{(1)}\mathbb{X}+b_{s}^{(1)})\right\}_{s=1}^{l}= g_1(\mathbf{U}_g^{(1)}\mathbf{a}_g^{(0)} + \mathbf{b}^{(1)}_g ).
		\end{equation}
		Note that $\mathbf{a}_g^{(0)}\in \mathbb{R}^{n\times k}$ and  $\mathbf{a}_g^{(1)}\in \mathbb{R}^{n\times l}$. If more than one hidden layer is assumed, $\mathbf{a}_g^{(1)}$ maps to the next hidden layer, otherwise it maps to the output layer.
		
		Consider for a moment balancing net with  one hidden layer. Then, the  forward propagation can be simply put together as follows:
		
		\textbf{Balancing net:}
		\begin{align}
			\mathbf{a}_g^{(0)}&=\mathbb{X}& (\text{Input Layer})\\
			\mathbf{a}_g^{(1)} &= g_1(\mathbf{U}_g^{(1)}\mathbf{a}_g^{(0)} + \mathbf{b}^{(1)}_g )& (\text{Hidden Layer})\\
			{a}_g^{(2)} &= g_2(\mathbf{u}_g^{(2)}\mathbf{a}_g^{(1)} + b^{(2)}_g) &(\text{Outer layer})
		\end{align}
		where $g_1$ and $g_2$ are activation functions, $\mathbf{U}_g^{(1)}$ represents the weights applied to the data from the input layer as it is mapped to the hidden layer, and $\mathbf{b}^{(1)}_g$ represents the biases added to these weights. Moreover, $\mathbf{u}_g^{(2)}$ denotes weight-multiplying information (data) from the hidden layer to the output layer, and $b^{(2)}_g$ denotes the bias added to it. These are weights and biases for balancing the net\footnote{The general forward propagation, when many hidden layers are involved, is given in Algorithm \ref{algorithm2}.}. The vector ${a}_g^{(2)}$ in the feedforward denotes the pseudo propensity score $g(\mathbb{X})$ in Equation \ref{outconereg}. This vector from the outer layer of the balancing net is then combined with $\mathbb{X}$ and $T$, i.e., $[\mathbb{X},T,{a}_g^{(2)}]$, and fed into the outcome net.
		
		Once the forward propagation in the balancing net is done, then the result ${a}_g^{(2)}$ from its outer layer will be combined with treatment and pretreatment variable as inputs for outcome net. These inputs  are then fed into the first hidden layer of outcome net. This step can be expressed as follows:
		\begin{equation}
			\left[\mathbb{X},T,{a}_g^{(2)}\right] \to \left\{f_1\left(\sum_{j=1}^{k}\beta_{j}^{(s,1)}X_j+\beta_{k+1}^{(s,1)}T+\beta_{k+2}^{(s,1)}{a}_g^{(2)}+b^{(s,1)}\right)\right\}_{s=1}^{l^{\prime}}, \label{fedoutcomenet}
		\end{equation}
		where $l^{\prime}$ denotes number neurons on the first hidden layer, and $f_1$ is activation function used on the first hidden layer. Following similar steps as in balancing net, one can denote  $\mathbf{u}^{(s,1)}=\left(\beta_{1}^{(s,1)},\beta_{2}^{(s,1)},\cdots, \beta_{k+2}^{(s,1)}\right)^{\top}$, $\mathbf{U}_f^{(1)}=\left\{\mathbf{u}^{(s,1)}\right\}_{s=1}^{l^{\prime}}$ and $\mathbf{b}^{(1)}_f=\left\{b^{(s,1)}\right\}_{s=1}^{l^{\prime}}$. These parameters are used during feedforward of input vectors of the outcome net. The outcome net obtains $\mathbb{X}$ and $T$ from the data at hand, while it obtains the pseudo propensity $a_g^{(2)}$ from the balancing net. Hence it introduces an extra dimension to the weights of which the score $\mathbf{u}_{\ast}^{(1)}=\left\{\beta_{s,k+2}\right\}_{s=1}^{l^{\prime}}$ is used to connect $a_g^{(2)}$ with the nodes of the first hidden layer of the outcome net.

		Similar to the balancing net, the outcome net can be designed using standard feedforward architecture. Unlike the balancing net, it has a supervised target, which is the outcome vector $Y$. Therefore, it incorporates a supervised loss function during the later stages of optimization. The feedforward architecture of the outcome net, similar to that of the balancing net, is described below:
		
		\textbf{Outcome Net:}
		\begin{align}
			\mathbf{a}_f^{(0)}&=[\mathbb{X},T,{a}_g^{(2)}]& \label{inputlayer} (\text{Input Layer})\\
			\mathbf{a}_f^{(1)} &= f_1( \mathbf{U}_f^{(1)}\mathbf{a}_f^{(0)} + \mathbf{b}^{(1)}_f )& (\text{Hidden Layer})\\
			\hat{Y} &= f_2(\mathbf{u}_f^{(2)}\mathbf{a}_f^{(1)} + b^{(2)}_f)&(\text{Output layer})
		\end{align}
		where $f_1$ and $f_2$ are activation functions, $\mathbf{U}_f^{(1)},\mathbf{b}^{(1)}_f,\mathbf{u}_f^{(2)},b^{(2)}_f$ are weights and biases of the outcome net. Weights and biases of balancing nets and outcome nets jointly form the weights and biases of self-balancing neural networks. These weights and biases can be initialized from Gaussian distribution \citep{glorot2010understanding}, like any other neural network weight initialization.

		Additionally, both balancing and outcome nets use the SiLU activation function suggested by \citet{elfwing2018sigmoid} or GELU, which was introduced by \citet{hendrycks2016gaussian} for the hidden layers. While linear activation is used at the outer layer of the balancing net and the output layer for the outcome net. Besides these activation functions, one can use the SWish activation function, which is similar to SILU and was independently discovered by \citet{ramachandran2017searching}, instead of SILU. The choice between Swish and SILU will depend on whether one is using PyTorch or TensorFlow to code the Sbnet model.
		
		Once architecture of a self-balancing neural network is proposed, the parameters of this model will be updated using backpropagation. The optimal parameters are reached when the model is fully trained. In the following subsection, the optimization stage is presented. Mainly backpropagation is presented in the subsection.
		
		\subsection{Optimization Stage}
		\label{optimizationstage}
		The forward propagation process uses the predefined architecture of the Sbnet, along with its weights and biases, to estimate the model's outcome. For final estimation, one needs to obtain optimal weights and biases. Gradients will backpropagate to update the Sbnet, after which feedforward will take place. These processes will continue until they reach optimal values, with the goal of achieving optimal weights and biases. The gradient backpropagation process follows the general principle of the backpropagation algorithm \citep{rumelhart1986learning, rumelhart1986learningmore}. 
		
		After completing forward propagation, gradient backpropagation will be done based on the loss incurred in estimating the outcome $Y$. This backpropagation will be performed to update the parameters of the Sbnet model. The loss function $\mathcal{L}$ for Sbnet follows the piecewise expression of the outcome variable $Y$. It is given as follows:
		\begin{equation}
			\mathcal{L}=\alpha\mathcal{L}_1+(1-\alpha)\mathcal{L}_0, \label{sbnetloss}
		\end{equation}
		where $\mathcal{L}_1$ and $\mathcal{L}_0$ are losses when treatment $T=1$ and $T=0$ respectively, and   $0<\alpha<1$ is the weighting parameter for the loses. Similar to any other neural network model, these losses can be chosen to be mean squared losses, mean absolute losses or hubber losses or others depending on how the losses works during training. 
		
		During backpropagation, gradient computation starts at the output layer. Based on the loss function gradient, the output layer is computed as follows:
		\begin{equation}
			\boldsymbol{g}=\nabla_{\hat{Y}}\mathcal{L} \label{initialgradient}
		\end{equation}
		Then, gradient backpropagation through the architecture obeys the chain rule. Start with a gradient on the output layer and backpropagate to the hidden layer closest to the output:
		
		$$\boldsymbol{\delta}^{(2)}_f=\nabla_{\mathbf{a}_f^{(2)}}\mathcal{L}= \boldsymbol{g}\odot f_2^{\prime}\left(\mathbf{u}_f^{(2)}\mathbf{a}_f^{(1)} + b^{(2)}_f\right)$$
		where $f_2^{\prime}$ represents differentiation of $f_2$. The notation $\odot$ is known as the Hadamard product, where it represents element-wise multiplication. Moreover, $\boldsymbol{\delta}^{(2)}_f$ is the gradient based on which the inner gradients are computed with respect to weight and bias and used to update weights and bias:
		\begin{align}
			\mathbf{b}_f^{(2)}&=\mathbf{b}_f^{(2)}- \lambda \boldsymbol{\delta}^{(2)}_f\\
			\mathbf{U}_f^{(2)}&=\mathbf{U}_f^{(2)}- \lambda \boldsymbol{\delta}^{(2)}_f \mathbf{a}^{(1)\top}_f
		\end{align}
		where $\lambda$ is the learning rate. Then with respect to the previous hidden layer
		$$\boldsymbol{\delta}^{(1)}_f= (\mathbf{U}^{(2)\top}_f\boldsymbol{\delta}^{(2)}_f)\odot f_1^{\prime}\left(\mathbf{U}_f^{(1)}\mathbf{a}_f^{(0)} + \mathbf{b}^{(1)}_f\right)$$
		this gradient is used to update $\mathbf{U}^{(1)}_f$  and $\mathbf{b}^{(1)}_f$ :
		\begin{align}
			\mathbf{b}_f^{(1)}&=\mathbf{b}_f^{(1)}- \lambda \boldsymbol{\delta}^{(1)}_f\\
			\mathbf{U}_f^{(1)}&=\mathbf{U}_f^{(1)}- \lambda \boldsymbol{\delta}^{(1)}_f\mathbf{a}^{(0)\top}_f
		\end{align}
		
		The weight $\mathbf{U}_f^{(1)}$ is a parameter connected to input $\mathbf{a}_f^{(0)}$, which contains pretreatment variables $\mathbb{X}$, treatment $T$, and the vector obtained from the outer layer of the balancing net ${a}_g^{(2)}$. Unlike $\mathbf{a}_g^{(0)}$ for balancing net, $\mathbf{a}_f^{(0)}$ becomes dynamic due to the nature of ${a}_g^{(2)}$. This component is not predefined as pretreatment variables $\mathbb{X}$ and treatment vectors $T$. Rather, we are required to do further optimization for the nested balancing net to yield this vector to obtain optimal ${a}_g^{(2)}$. Hence, it necessitates further computation of the gradient and backpropagation to achieve its optimality.
		
		\begin{algorithm}[h!]
			\caption{Self Balancing Neural Network Algorithm}
			\label{algorithm2}
			\begin{algorithmic}[1]
				\State \textbf{Initialize:} Sbnet Weight Matrices $\mathbf{U}$ and bias vectors $\mathbf{b}$
				\State \textbf{Input:} Pre-treatment variables $\mathbb{X}$, Treatment $T$ and Outcome $Y$
				
				\State 	$\mathbf{a}_g^{(0)}=\mathbb{X}$ \hfill \textcolor{gray}{// Input for balancing net}
				\For{in epochs}
				\Statex \textcolor{gray}{~~~~Forward propagation}
				
				\For{$h=1,\ldots,d$} \hfill \textcolor{gray}{// $d$ hidden layers are considered}
				\State $\mathbf{a}_g^{(h)} = g_h(\mathbf{U}_g^{(h)}\mathbf{a}_g^{(h-1)} + \mathbf{b}^{(h)}_g )$ \hfill \textcolor{gray}{// Hidden layers for balancing net}
				
				\EndFor
				\State ${a}_g^{(d+1)} = g_{d+1}(\mathbf{u}_g^{(d+1)}\mathbf{a}_g^{(d)} + b^{(d+1)}_g)$ \hfill \textcolor{gray}{// Pseudo propensity score}
				\State $\mathbf{a}_f^{(0)}=[\mathbb{X},T,{a}_g^{(d+1)}]$ \hfill\textcolor{gray}{// Input for outcome net}
				\For{$h=1,\ldots,m$} \hfill \textcolor{gray}{// $m$ hidden layers are considered}
				\State $\mathbf{a}_f^{(h)} = f_h( \mathbf{U}_f^{(h)}\mathbf{a}_f^{(h-1)} + \mathbf{b}^{(h)}_f )$ \hfill \textcolor{gray}{// Hidden layers for outcome net}
				\EndFor
				\State $\hat{Y} = f_{m+1}(\mathbf{u}_f^{(m+1)}\mathbf{a}_f^{(m)} + b^{(m+1)}_f)$ \hfill \textcolor{gray}{// Estimated outcome}
				\Statex
				\Statex \textcolor{gray}{~~~~Back propagation}
				\State $\boldsymbol{g}=\nabla_{\hat{Y}}\mathcal{L}$  \hfill \textcolor{gray}{// Gradient due to estimated outcome}
				\For{$h=m,\ldots,1$} \hfill \textcolor{gray}{// Backpropagation loop for outcome net}
				\State $\boldsymbol{\delta}^{(h+1)}_f=\nabla_{\mathbf{a}_f^{(h+1)}}\mathcal{L}= \boldsymbol{g}\odot f_{h+1}^{\prime}\left(\mathbf{U}_f^{(h+1)}\mathbf{a}_f^{(h)} + \mathbf{b}^{(h+1)}_f \right)$
				\State $	\mathbf{b}_f^{(h+1)}=\mathbf{b}_f^{(h+1)}- \lambda \boldsymbol{\delta}_f^{(h+1)}$
				\State $
				\mathbf{U}_f^{(h+1)}=\mathbf{U}_f^{(h+1)}- \lambda \boldsymbol{\delta}_f^{(h+1)}\mathbf{a}^{(h)\top}_f$
				\State $\mathbf{g}=\mathbf{U}^{(h+1)\top}_f\boldsymbol{\delta}^{(h+1)}_f$
				\EndFor
				\State $\mathbf{d}=\mathbf{u}^{(1)}_{\ast}\boldsymbol{\delta}_f^{(1)}$ \hfill \textcolor{gray}{// Gradient due to outer layer of balancing net}
				\For{$h=d,\ldots,1$} \hfill \textcolor{gray}{// Backpropagation loop for balancing net}
				\State $\boldsymbol{\delta}^{(h+1)}_g= \boldsymbol{d}\odot g_{h+1}^{\prime}\left(\mathbf{U}_g^{(h+1)}\mathbf{a}_g^{(h)} + \mathbf{b}^{(h+1)}_g\right)$
				\State $	\mathbf{b}_g^{(h+1)}=\mathbf{b}_g^{(h+1)}- \lambda \boldsymbol{\delta}^{(h+1)}$
				\State $
				\mathbf{U}_g^{(h+1)}=\mathbf{U}_g^{(h+1)}- \lambda \boldsymbol{\delta}^{(h+1)}\mathbf{a}^{(h)\top}_g$
				\State $\mathbf{d}=\mathbf{U}^{(h+1)\top}_g\boldsymbol{\delta}^{(h+1)}_g$
				\EndFor
				\EndFor

			\end{algorithmic}
		\end{algorithm}

		Unlike the gradient computation done at the output layer of the outcome net in Equation \ref{initialgradient}, the gradient for the outer layer of the balancing net will not be done by differentiating the loss function with respect to ${a}_g^{(2)}$. Instead, we obtain the gradient by sharing gradients on the input layers of the outcome net. This form of optimization, which involves interaction with its environment, is similar to other reinforcement learning methods. Simply put, the shared gradient represents the reward that balancing nets receive from outcome nets as a result of their feedforward behavior.
		
		As stated earlier, the weight $\mathbf{U}_f^{(1)}$  denotes weights connecting input vectors  $\mathbf{a}_f^{(0)}=[\mathbb{X},T,{a}_g^{(2)}]$ to hidden layers. Specifically, $\mathbf{u}^{(1)}_{\ast}$ denote weight from $\mathbf{U}_f^{(1)}$ corresponding to ${a}_g^{(2)}$. Then the gradient with respect to the outer layer of balancing net  is obtained as follows:
		
		$$\boldsymbol{\delta}^{(2)}_g= (\mathbf{u}^{(1)}_{\ast}\boldsymbol{\delta}^{(1)}_f)\odot f_1^{\prime}\left(\mathbf{U}_f^{(1)}\mathbf{a}_f^{(0)} + \mathbf{b}^{(1)}_f\right)$$
		which is equivalent to gradient computed in \ref{initialgradient}, for balancing net. Similar to backpropagation in the outcome net, the gradient will follow chain rule for the backpropagation. Hence, gradient backpropagate to the next hidden layer of balancing net:
		
		$$\boldsymbol{\delta}^{(1)}_g= (\mathbf{U}^{(2)\top}_g\boldsymbol{\delta}^{(2)}_g)\odot g_2^{\prime}\left(\mathbf{u}_g^{(2)}\mathbf{a}_g^{(1)} + b^{(2)}_g\right).$$

		Similar to previous steps weights and biases in balancing net will be updated as follows:
		\begin{align}
			\mathbf{b}_g^{(2)}&=\mathbf{b}_g^{(2)}- \lambda \boldsymbol{\delta}^{(2)}_g\\
			\mathbf{U}_g^{(2)}&=\mathbf{U}_g^{(2)}- \lambda \boldsymbol{\delta}^{(2)}_g\mathbf{a}^{(1)\top}_g
		\end{align}
		and 
		\begin{align}
			\mathbf{b}_g^{(1)}&=\mathbf{b}_g^{(1)}- \lambda \boldsymbol{\delta}^{(1)}_g\\
			\mathbf{U}_g^{(1)}&=\mathbf{U}_g^{(1)}- \lambda \boldsymbol{\delta}^{(1)}_g\mathbf{a}^{(0)\top}_g.
		\end{align}
		
		Once the model is fully trained, the vector $a_g^{(2)}$ obtained from the balancing net is retained as the optimal pseudo propensity score. 
		
		Generally, this sketch illustrates the process of backpropagation within a self-balancing net. In Algorithm \ref{algorithm2}, both the feedforward and backpropagation processes of Sbnet are presented. In the following subsection, we present how to assess the optimality of the pseudo propensity score.
		
		\subsection{Assessing Behavior of Pseudo Propensity Score}
		Models that use propensity scores typically depend on treatment models to generate predictive probabilities for estimating those scores. These models are established by relying on the minimization loss functions of the classification models used.
		
		On the other hand, pseudo propensity scores obtained from balancing nets in Sbnet architecture do not have a separate loss function as propensity score models do. Instead, they form an essential component of the Sbnet architecture and undergo optimization in tandem with the outcome network. Therefore, we need to determine how to assess the convergence of the parameters for pseudo propensity scores, specifically the weights and biases. 
		
		For every update during the training process, let $\theta(g_{(c)})$ be the parameters of the current pseudo propensity score and $\theta(g_{(p)})$ be the parameters (i.e., weights and biases) of the previous pseudo propensity score. Here $g_{(c)}$ and $g_{(p)}$ are simplified expressions for $g_{(c)}(\mathbb{X})$ and $g_{(p)}(\mathbb{X})$. To establish convergence of the pseudo propensity score, we rely on whether the consecutive parameters $\theta(g_{(p)})$ and $\theta(g_{(c)})$ become closer and closer each time when the Sbnet becomes fully trained.
		
		If the distribution functions $F_{\theta(g_{(p)})}$ and $F_{\theta(g_{(c)})}$ are known, then using a strong distance measure $\mu$ between them, one can establish whether the parameters are discriminable or not. For $\mu(F_{\theta(g_{(p)})},F_{\theta(g_{(c)})})= 0$, the parameters $\theta(g_{(p)})$ and $\theta(g_{(c)})$ are not discriminable. However, the distribution functions are intractable for the pseudo propensity scores obtained during training; specifically, the balancing net within the Sbnet architecture, which generates these pseudo propensity scores, is a black box model.
		
		To handle this problem, we can take advantage of the sampler, as the black box models are data-generating processes \citep{yatracos2024edi}. In our case, a pseudo propensity score is generated each time and fed into the outcome net; then, during the balancing net, which generates pseudo propensity score updates, its parameters continue to generate pseudo propensity scores. Thus the distance $\mu(F_{\theta(g_{(p)})},F_{\theta(g_{(c)})})$ can be estimated by $\mu(\hat{F}(g_{(p)}),\hat{F}(g_{(c)})$. The resulting  distance $\mu(\hat{F}(g_{(p)}),\hat{F}(g_{(c)}))$ become a random variable. Thus one can obtain a $p-value$ to see how large $\mu(F_{\theta(g_{(p)})},F_{\theta(g_{(c)})})$ is.
		
		In our case, we do not rely on a single or one-time distance measure to establish convergence of pseudo propensity score generation. Instead, we want to see the pattern of this distance, whether it is becoming closer and closer or not. When they continuously become closer and closer, we conclude that the parameters of consecutive pseudo propensity scores become indiscriminable. And hence, pseudo propensity score generation done by balancing net becomes convergent. So, the distance between the pseudo propensity scores from one training step to the next should get smaller and smaller, $\mu(\hat{F}(g_{(p)}),\hat{F}(g_{(c)}))< \epsilon$, and hence the parameters lie in a small neighbourhood $N_{\epsilon}$.
		
		To test that the consecutive parameters of balancing nets generating pseudo propensity scores lie in a small neighbourhood, the distance $\mu$ can be computed using the Wasserstein distance, symmetric KL divergence, or other distance measures. Using these measures, one can determine the optimality of the pseudo propensity score.
		
		Moreover, since the pseudo propensity score works as a balancing score within the Sbnet architecture, one can directly examine how stable these scores are by looking at the loss $\mathcal{L}$ over time. When loss reduction is consistent, parameters of Sbnet are beginning to converge. This implies the convergence of the balancing net's parameters nested inside Sbnet, thereby establishing the optimality of the pseudo propensity score.

		\subsection{Sbnet with Multiple Pseudo Propensity Scores}
		The formulation of a self-balancing neural network allows us to reframe the problem as a structural improvement, enabling the neural network to better estimate the average treatment effect. Such formulations therefore do not limit us to having a single balancing net. Hence, one may be interested in more than one balancing net so that they jointly improve the result of estimation.
		
		At this point, let us denote the functional form of Sbnet whose outcome net has no hidden layers. Then it can be expressed as a standard outcome regression model:
		$$Y=\gamma T+\mathbb{X}\boldsymbol{\beta}+\vartheta g(\mathbb{X}), $$
		where $\gamma,\boldsymbol{\beta}$ and $\vartheta$ are parameters of the model. For some orthogonal $g_i(\mathbb{X})$, $i=1,2,\ldots,m$, such that $\sum_i\omega_i g_i(\mathbb{X}) =g(\mathbb{X})$ where $\omega_i \in \mathbb{R}$, one can easily obtain the following expression:
		$$Y=\gamma T+\mathbb{X}\boldsymbol{\beta}+\vartheta \sum_{i=1}^m\omega_i g_i(\mathbb{X}). $$
		
		Since orthogonal components provide distinct information, one can call them diversified vectors. The resulting $g_i(\mathbb{X})$ can be considered an estimate of the pseudo propensity score. Based on these diversified models, a set of pseudo propensity score estimations will be collected, just like in ensemble learning. Using similar intuition, one can establish the notion of multiple pseudo propensity scores in Sbnet using multiple balancing nets.
		
		Suppose that there are $m$ number of balancing nets involved in generating pseudo propensity scores; then the forward propagation inside outcome net expressed in \ref{fedoutcomenet} for Sbnet with a single balancing net will be improved as follows:
		\begin{equation}
			\resizebox{0.75\textwidth}{!}{$	\left[\mathbb{X},T,{a}_{g_1}^{(2)},{a}_{g_2}^{(2)},\cdots,{a}_{g_m}^{(2)}\right] \to \left\{f_1\left(\sum_{j=1}^{k}\beta_{j}^{(s,1)}X_j+\beta_{k+1}^{(s,1)}T+\sum_{i=k+2}^{k+m+1}\beta_{i}^{(s,1)}{a}_{g_{i-(k+1)}}^{(2)}+b^{(s,1)}\right)\right\}_{s=1}^{l^{\prime\prime}}$}.
		\end{equation}
		Then the backward propagation algorithm will easily improve by sharing gradients into multiple balancing nets. Each balancing net updates their parameters based on the gradients.
		
		\begin{figure}[h]
			\centering
			\includegraphics[width=0.5\textwidth]{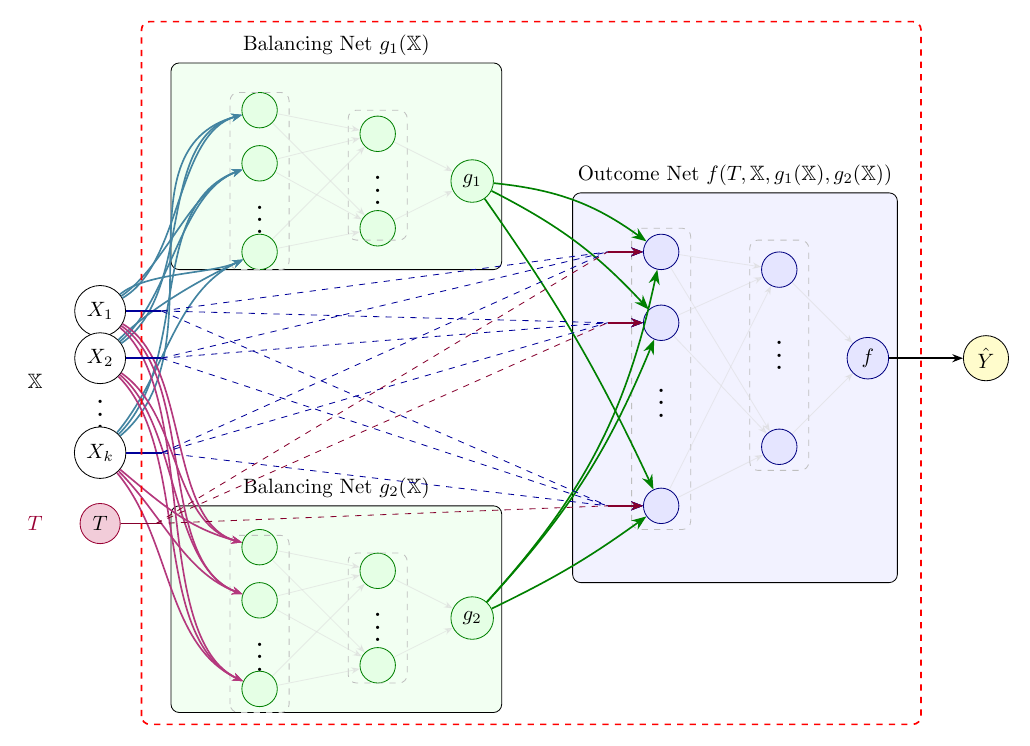}
			\caption{Self Balancing Neural Network architecture with two pseudo propensity scores. Both arrows with broken lines and without broken lines indicate the forward propagation from input to hidden layer. The broken lines are used only for the sake of visibility.}
			\label{Dbnet}
		\end{figure}

		Since the goal is to improve the estimation by incorporating multiple balancing nets, they should be diversified so that they provide different information for outcome nets. To attain this diversification, one can rely on varying numbers of hidden layers and nodes or use different activation functions for each balancing net. Distinct activation functions are believed to be the source of diversification; for example, when ensemble learning is involved, \citep{Goodfellow-et-al-2016}. Diversification can also be achieved by modelling subsets of the pretreatment variables $\mathbb{X}$ with each balancing net. This paper relies on distinct activation functions as the source of diversification.
		
		Consider the case where $m=2$ balancing nets are considered to estimate pseudo propensity scores. This type of self-balancing architecture can be depicted as in Figure \ref{Dbnet}. In comparison to the self-balancing net described in Figure \ref{Sbnet}, it involves one additional balancing net. This requires structural improvements to the input layer of the outcome net to handle both balancing nets. 
		
		\begin{equation}
			\mathbf{a}_f^{(0)}=\left[\mathbb{X},T,{a}_{g_1}^{(2)},{a}_{g_2}^{(2)}\right]
		\end{equation}
		
		In this context, ${a}_{g_1}^{(2)}$ and ${a}_{g_2}^{(2)}$ refer to vectors that are derived from the outer layers of balancing nets during the forward propagation process. As a result, during backpropagation, there will be an extra balancing net that requires gradient to update its weights and biases. Accordingly, the Algorithm \ref{algorithm2} given for a standard self-balancing net can easily be improved.

		In this scenario, conditional independence in Assumption \ref{assmp2} is defined based on joint pseudo propensity scores:
		$$Y(t) \perp T | g_1(\mathbb{X}),g_2(\mathbb{X}), \quad \text{where } t=0,1. $$
		
		Both $g_1(\mathbb{X})$ and $g_2(\mathbb{X})$ are integrated inside Sbnet similar to the previous case. Hence the optimal structure of them is attained when the loss function in Equation \ref{sbnetloss} attains a minimum. Moreover, to detect the stability of these pseudo propensity scores, one can rely on the pattern of their consecutive distance measures. One will perform this procedure separately for each of them. Then one can conclude the stability of the pseudo propensity scores and further support the optimality of the resulting measures. $g_1(\mathbb{X})$ and $g_2(\mathbb{X})$ from this Sbnet architecture.	
		
		Choosing the optimal number of balancing nets for pseudo propensity scores can be easily considered similar to the standard way of finding optimal architecture in neural network models. The following sections provide simulation studies and real-world applications of the proposed setting. 
		
		\section{Simulation}
		\label{sec4}
		Simulation is performed to assess the performance of the proposed method. As benchmark models, a total of nine (9) existing methods are considered to estimate the average treatment effect and be compared with the proposed method under various simulation settings. Two of these methods are based on the backdoor method in DoWhy, using the inverse propensity weight and the propensity score model. The remaining methods are obtained from machine learning models: These methods include XGBTRegressor, LRSRegressor, MLPTRegressor, Causal Forest DML, Kernel DML, TARNet, and Dragonet.
		
		The following subsections present the data generation process and the estimated ATE results. Data sets with a sample size of 1000 are generated 100 times. Each time, the data are split into $80\%$ training and $20\%$ testing. All methods, except the DoWhy method, predict ATE on test data. For DoWhy methods, the test data is refitted since they are not predictive models. To obtain bootstrap standard deviations, the models are refitted for each resampled data set across 50 resamplings of the generated data.

		\subsection{Data Generation Process}
		This simulation study uses three different treatment-generating mechanisms to compare the proposed Sbnet model with existing methods. The first case is when there is no strong confounder. All models are expected to yield good results, i.e., unbiased, in this scenario. In Case 2 and Case 3, treatment assignments are generated assuming the presence of a strong confounder and an absolute value function being involved.
		
		As a function of treatment and predictors in this simulation setup, the outcome model is generated as follows:
		\begin{equation}
			Y =  8 +2T+ (X_1-0.3)^2 + 0.5X_2^3+ 0.6(X_3X_4) + \sin(-0.5(X_5+X_6)) + \epsilon,
		\end{equation}
		where $\epsilon \sim N(0,0.01)$ and treatment is generated as follows:
		\begin{align*}
			T&\sim binom(1,e(\mathbb{X}))\\
			\text{Case 1: } 	e(\mathbb{X})&=\frac{1}{1+\exp{(X_1-X_7-0.5+\epsilon_1)}},
			\text{ Case 2: } 	e(\mathbb{X})=\frac{1}{1+\exp{(4(X_1-0.5)-X_7+\epsilon_2)}} \quad \text{and}\\
			\text{Case 3: } 	e(\mathbb{X})&=\frac{1}{1+\exp{(|4(X_1-0.5)-X_7|+\epsilon_3)}}
		\end{align*}
		where $\epsilon_1\sim N(0,0.01),\epsilon_2,\epsilon_3 \sim N(0,0.1)$. For all cases $\mathbb{X}=2\Phi(W)-1$, where $W_{n \times 7}\sim \mathcal{N}(\boldsymbol{0},\Sigma)$, $\Sigma= \{\sigma_{j,l}\}_{j,l=1}^7$ and $\sigma_{j,l}=0.5^{|j-l|}$. 
		
		The following subsection presents the estimated average treatment effect using the proposed Sbnet method, based on the data generation processes mentioned above. Moreover, nine existing methods are used to estimate ATEs for the same datasets, and these methods are compared to the proposed method's performance.

		\subsection{Results}

\begin{figure*}[h]
			\centering
			
			\begin{subfigure}[t]{0.475\textwidth}
				\centering
				\includegraphics[width=\textwidth]{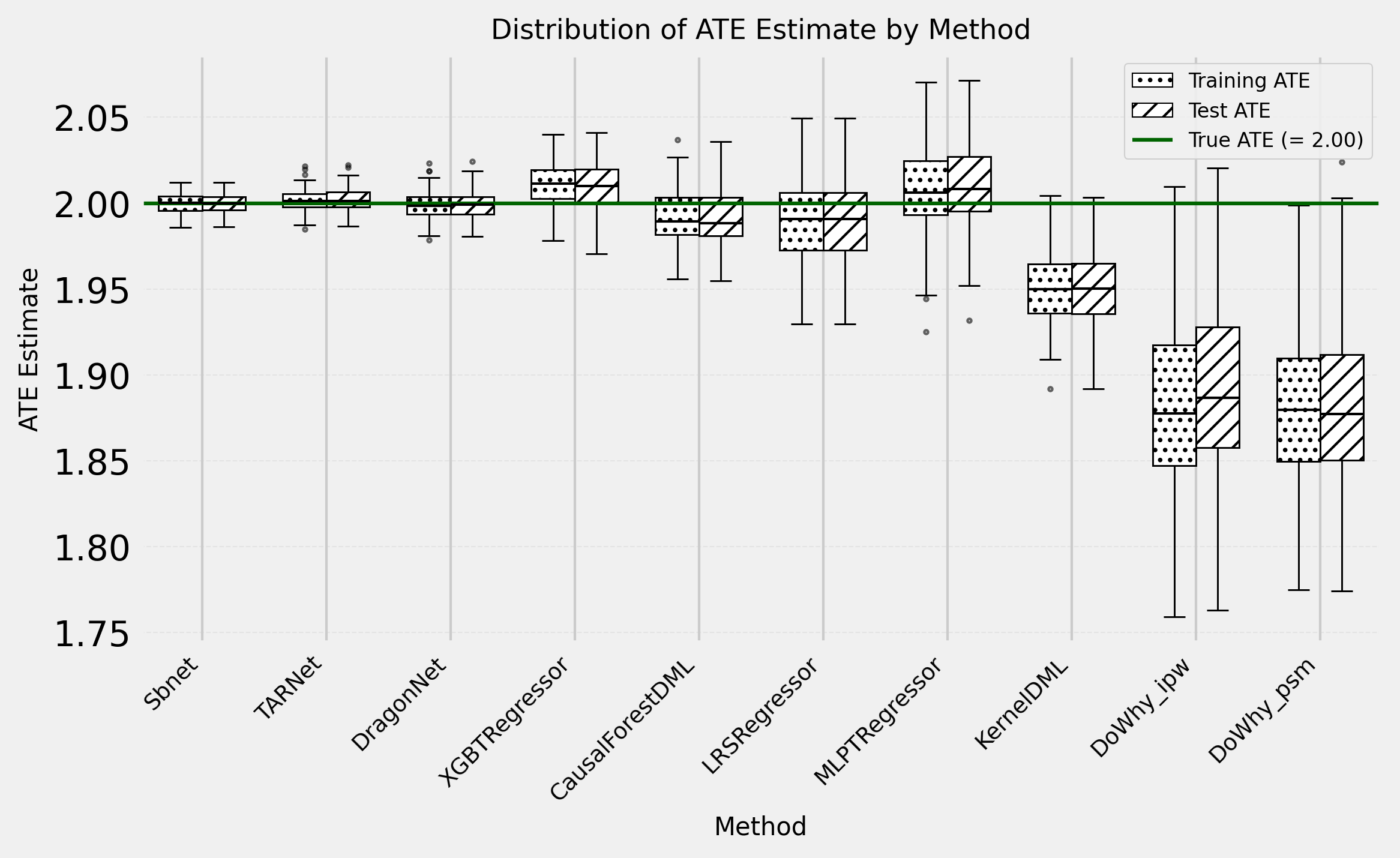}
				\caption{Case 1}
				\label{boxplot:figure1}
			\end{subfigure}
			\hfill  
			\begin{subfigure}[t]{0.475\textwidth}
				\centering
				\includegraphics[width=\textwidth]{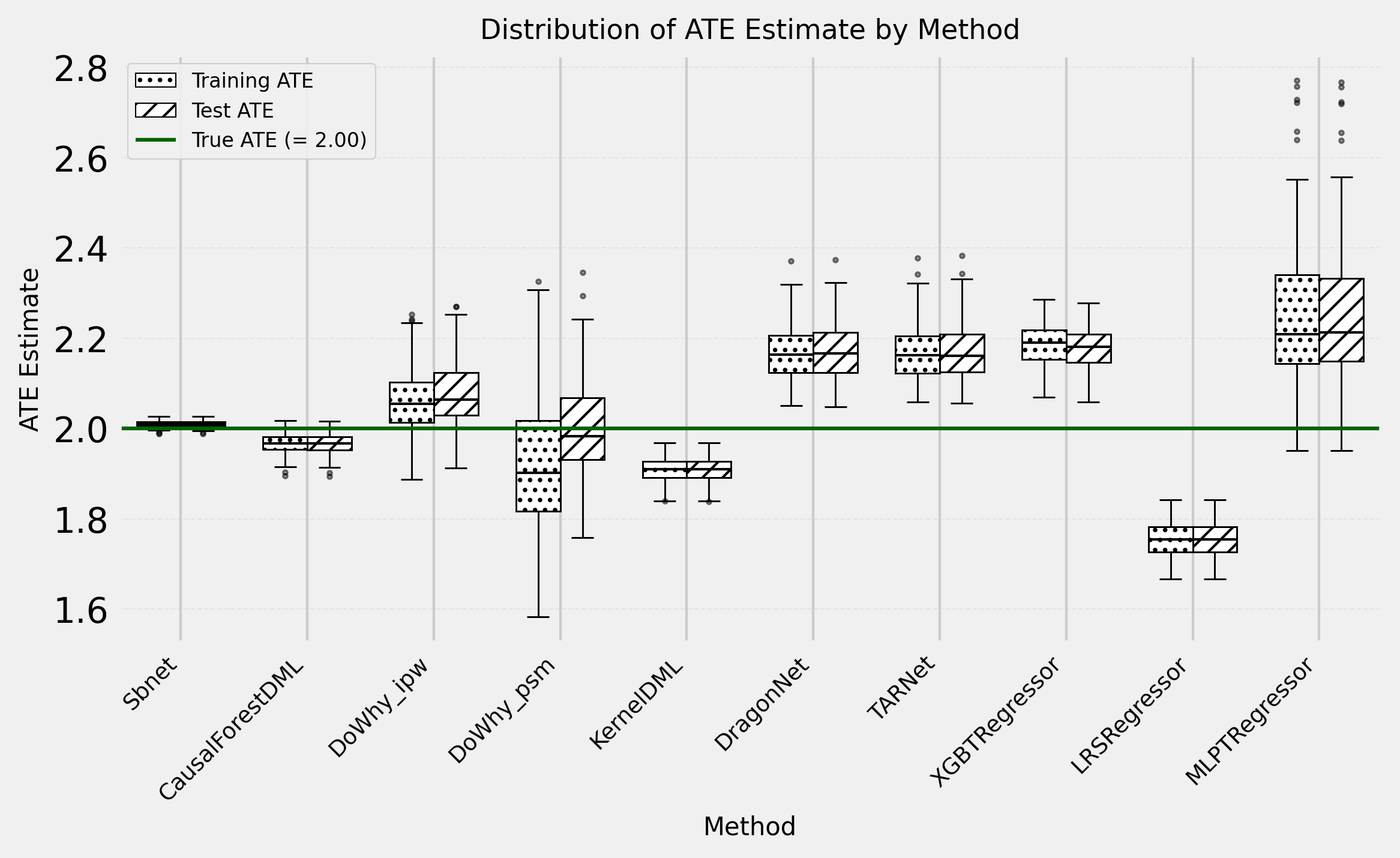}
				\caption{Case 2}
				\label{boxplot:figure2}
			\end{subfigure}
			\vspace{0.001cm} 
			\begin{subfigure}[t]{0.475\textwidth}
				\centering
				\includegraphics[width=\textwidth]{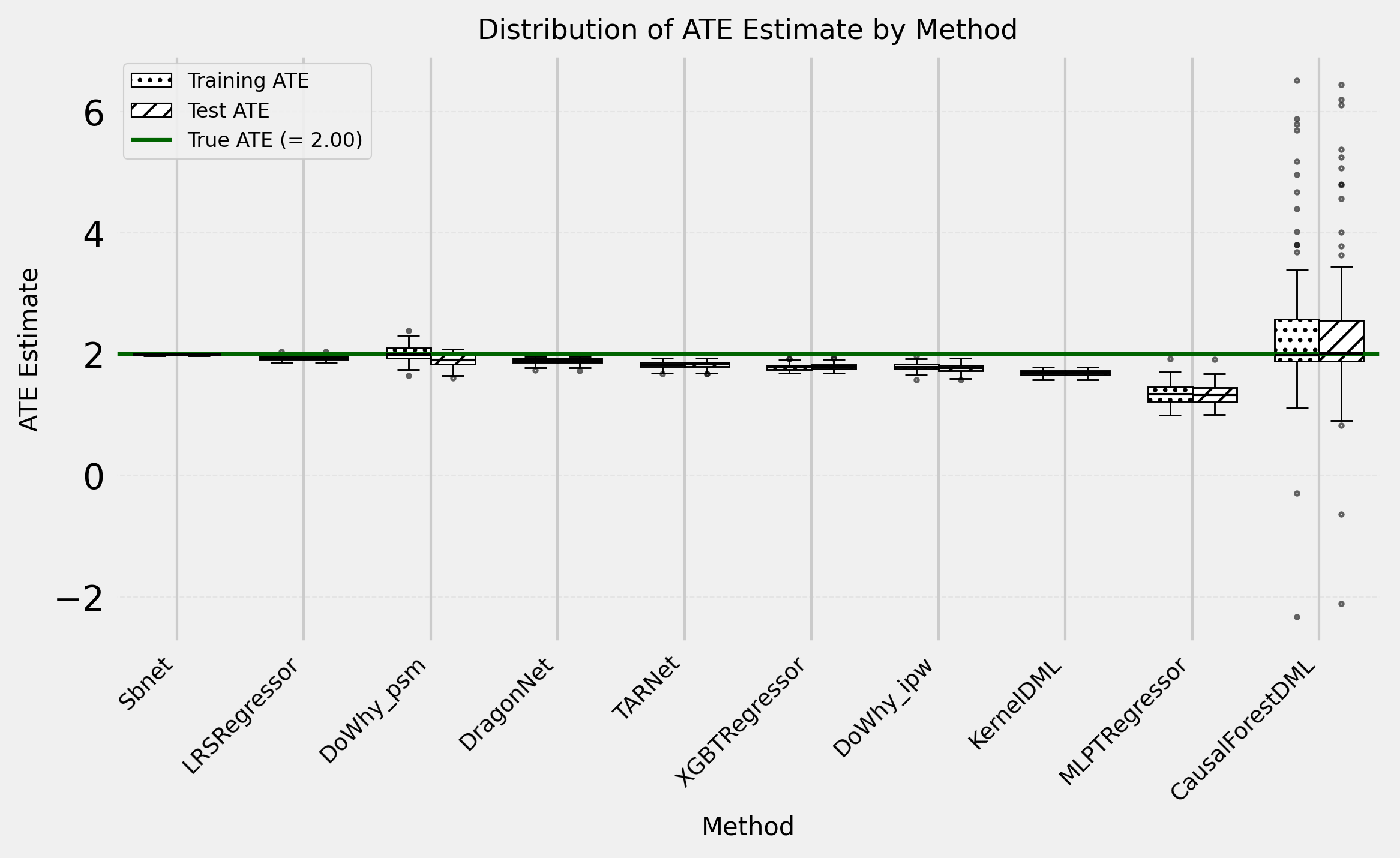}
				\caption{Case 3}
				\label{boxplot:figure3}
			\end{subfigure}
			\caption{Boxplots of the estimated average treatment effect distributions. For the proposed simulation setups, the distribution of the estimated ATE by each method is given.}
			\label{boxplot:mainfig}
		\end{figure*}

		The results from the simulation study show how well the new method and the exiting methods estimate the Average Treatment Effect (ATE) in various situations. In Figures \ref{boxplot:figure1} to \ref{boxplot:figure3}, the distribution of the estimated Average Treatment Effect (ATE) is provided using box plots which show how close the estimation of ATE is to the true ATE.

		For all cases,  Sbnet, whose balancing and outcome nets have one hidden layer with 128 nodes, fits well. For both nets, the SiLU activation function is at the hidden layer. Moreover, a linear activation function is used at outer layers. The proposed architecture is trained for 3000 epochs with a batch size of 64. During the training, to update the parameters of Sbnet, the loss function $\mathcal{L}$, which is computed from the mean squared loss functions $\mathcal{L}_0$ and $\mathcal{L}_1$ combined with equal weights $\alpha=0.5$, is minimized. The training stops early if the validation loss increases beyond the best validation loss for 50 consecutive epochs. Initially, the best validation loss is fixed at infinity, and a 0.0001 threshold is used to determine whether it shows improvement or not. The validation data is obtained from $10\%$ of the training data.
		
		For Case 1 shown in Figure \ref{boxplot:figure1}, almost all methods show results closer to true ATE. The Dowhy methods exhibit some deviation from the true Average Treatment Effect (ATE). For Cases 2 and 3, some methods show higher deviations from true ATE. The boxplots reveal that the nature of treatment models significantly influences existing methods.

		\begin{table}[h]
			\centering
			\centering
			\small  
			\setlength{\tabcolsep}{0.75mm}  
			\caption{Bias of average treatment effect estimates for the simulated data}
			
			\begin{tabular}{lrrrrrr}
				\toprule
				\multirow{3}{*}{\textbf{Method}} & \multicolumn{2}{c}{\textbf{Case 1}} & \multicolumn{2}{c}{\textbf{Case 2}} & \multicolumn{2}{c}{\textbf{Case 3}}\\ \cmidrule(lr){2-3} \cmidrule(lr){4-5}\cmidrule(lr){6-7}
				&\multicolumn{2}{c}{$\quad \quad\quad $\textbf{Bias (std)}} & \multicolumn{2}{c}{$\quad \quad\quad $\textbf{Bias (std)}} & \multicolumn{2}{c}{$\quad \quad\quad $ \textbf{Bias (std)}}\\
				&Training &Test  &Training &Test &Training &Test \\
				\midrule

				DoWhy PMS       &  -0.121 (0.034) &-0.118 (0.103) &  -0.086 (0.088) & 0.004 (0.258)& 0.012 (0.086) &  -0.093 (0.251)\\
				DoWhy IPW      &   -0.120 (0.023) &  -0.107 (0.094) & 0.061 (0.037) & 0.078 (0.140)& -0.211 (0.035)&  -0.231 (0.134)\\
				KernelDML       &   -0.049 (0.027) & -0.049 (0.027) & -0.092 (0.038) & -0.092 (0.038)&  -0.310 (0.097)&  -0.310 (0.098)\\
				CausalForestDML &  -0.008 (0.022) & -0.009 (0.023) &  -0.033 (0.032) &  -0.033 (0.032)& 0.349 (2.995) & 0.379 (3.129)\\
				DragonNet     & -0.001 (0.011) & -0.001 (0.013) & 0.171 (0.071) & 0.172  (0.076)& -0.108 (0.057) & -0.110 (0.060)\\
				TARNet        &  0.002 (0.010) & 0.002 (0.011) & 0.173  (0.066) & 0.174  (0.070)&-0.175 (0.060)& -0.176 (0.063)\\
				XGBTRegressor   &0.011 (0.010) & 0.009 (0.020) &0.185 (0.027) & 0.179 (0.040)& -0.215 (0.034) &  -0.208 (0.045)\\
				LRSRegressor   &   -0.009 (0.014) &  -0.009 (0.014) &  -0.245 (0.020) &  -0.245  (0.020)& -0.063 (0.018) &  -0.063 (0.018)\\
				MLPTRegressor   &   0.006 (0.020) & 0.009 (0.031) & 0.254 (0.144) & 0.256 (0.146)&  -0.658 (0.175) &  -0.669 (0.207)\\
				\textbf{Sbnet }   &\bfseries{0.000 (0.005)} & \bfseries{0.000 (0.005)}     &  \bfseries{0.011 (0.010)} & \bfseries{0.011 (0.011)}&\bfseries{-0.010
					(0.009)} & \bfseries{-0.010 (0.010)}\\
				
				\bottomrule 
			\end{tabular}
			\begin{tablenotes}
				\item \footnotesize Note: The value in the table displays the bias of estimated average treatment effect (ATE), along with the standard deviation obtained through bootstrapping. The estimated standard deviations are presented in the brackets.
			\end{tablenotes}
			\label{bigtable}
		\end{table}

		
		Moreover, the results of the simulations are presented in Table \ref{bigtable}. In the table, the bias of estimated average treatment effect and its standard deviation, for each cases, are reported. Standard deviations are computed based on 50 realizations (bootstrapping), and the values are reported in brackets.
		
		One can see that existing methods are highly influenced by the strength of confounding variables in the treatment model and also the functional form of the treatment model. For example, in Case 2, when there is a stronger confounding variable involved in the treatment model, Causal Forest DML shows better performance than other existing methods. However, in Case 3, when treatment models are more complex, Causal Forest DML performs worse compared to other existing methods. The standard deviation of its bias becomes inflated.  In general, Figures~\ref{boxplot:mainfig}(\subref{boxplot:figure1} - \subref{boxplot:figure3}), and Table \ref{bigtable} clearly show that the effectiveness of other methods varies depending on the treatment model.
		
		On the other hand, in all cases, estimated values of ATE obtained using the proposed Sbnet method shows the lowest bias than the existing methods. The Sbnet demonstrates sufficient flexibility and robustness to manage the covariates and functional form in the treatment model. The pseudo propensity score generated inside Sbnet did very well to balance covariates, and hence the resulting ATE became an unbiased estimate.

		\begin{table}[H]
			\centering
			\small  
			\setlength{\tabcolsep}{1mm}  
			\caption{Comparisons of Sbnet models' performances on the simulated data sets.}
				\begin{tabular}{llrrrr}
					\toprule
					\multirow{2}{*}{\textbf{Cases}}&\multirow{2}{*}{\textbf{Method}} & \multicolumn{2}{c}{\textbf{Bias (std)}} &\multicolumn{2}{c}{\textbf{Loss (std)}} \\ \cmidrule(lr){3-4} \cmidrule(lr){5-6} 
					&&Training &Test &Training &Test \\
					\midrule
					\multirow{2}{*}{\textbf{Case 1}}&Sbnet&0.000 (0.005)& 0.000 (0.005) &0.009 (0.004)&0.011 (0.005)\\
					&Sbnet (DPP)&0.000 (0.002)& 0.000 (0.003)& 0.002 (0.002)&0.002 (0.003)\\ 
					&&&&&\\
					\multirow{2}{*}{\textbf{Case 2}}&Sbnet& 0.011 (0.010)& 0.011 (0.011)&  0.010 (0.004)&0.012 (0.004)\\
					&Sbnet (DPP)&0.003 (0.006) &0.003 (0.006)& 0.002 (0.003)&0.003 (0.004)\\ 
					&&&&&\\
					\multirow{2}{*}{\textbf{Case 3}}&Sbnet& -0.010 (0.009)&-0.010 (0.010)& 0.013 (0.009)&0.014 (0.003)\\
					&Sbnet (DPP)&-0.002 (0.008) &-0.002 (0.008)& 0.004 (0.005)&0.006 (0.006)\\ 
					\bottomrule
				\end{tabular}
			
			\begin{tablenotes}
				\item \footnotesize Note: The value in the table displays the bias of estimated average treatment effect (ATE), and the loss between the true outcome $Y$ and the estimated outcome $\hat{Y}$, along with the standard deviation obtained through bootstrapping. Sbnet (DPP) refers to Sbnet with double pseudo propensity scores.
			\end{tablenotes}
			\label{sbnetvssbnetdpp}
		\end{table}

		Moreover, Sbnet with double pseudo propensity score (Sbnet DPP) is fitted. In this case, two balancing nets are involved, where the first balancing net uses the SiLU activation function and the other balancing net uses the GELU activation function. For both balancing nets, one hidden layer with 128 nodes is used, and Linear activation functions are used at the outer layer for both balancing nets. Then the result is fed into the outcome net. 
		
		Table \ref{sbnetvssbnetdpp} presents the result of comparisons of self-balancing neural network architectures. From the table, one can  see that self-balancing with a double pseudo propensity score shows better performance in estimating ATE. The loss and absolute bias of Sbnet with double pseudo propensity scores are lower than those of Sbnet with one pseudo propensity score. This indicates that multiple pseudo propensity scores can be utilized to improve the estimation process with Sbnet.

		\begin{figure*}[h]
			\centering
			
			\begin{subfigure}[b]{0.475\textwidth}
				\centering
				\includegraphics[width=\textwidth]{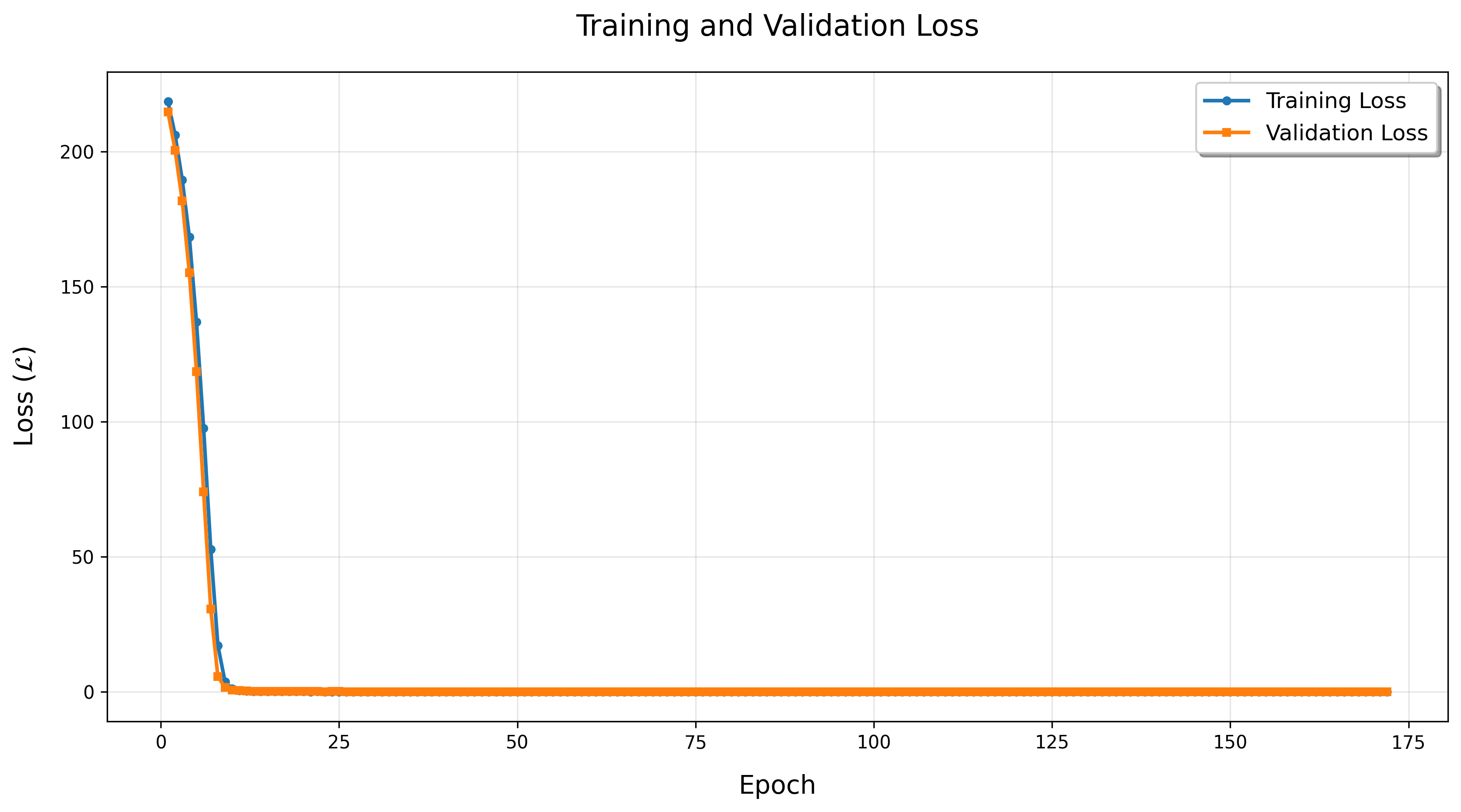}
				\caption{Loss}
				\label{case3sbnet:subfig1}
			\end{subfigure}
			\hfill  
			\begin{subfigure}[b]{0.475\textwidth}
				\centering
				\includegraphics[width=\textwidth]{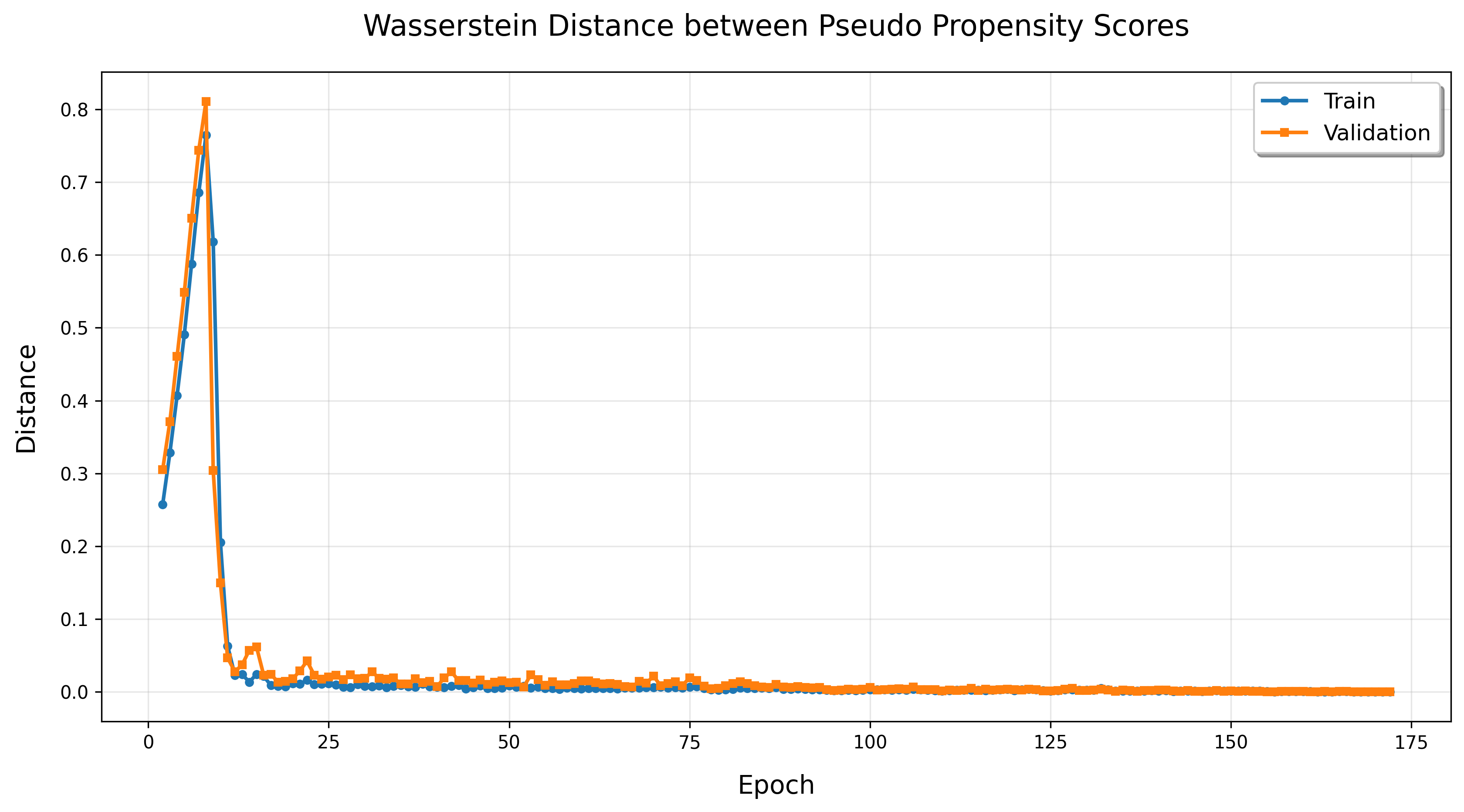}
				\caption{Wasserstein distance}
				\label{case3sbnet:subfig2}
			\end{subfigure}
			
			\caption{Pattern of loss in estimating outcome and Wasserstein distance between consecutive propensity scores (for Case 3). The loss is computed for all iterations, while the computation of Wasserstein distance starts at the second iteration.}
			\label{case3sbnnet:mainfig}
		\end{figure*}

		\begin{figure*}[h]
			\centering
			
			\begin{subfigure}[b]{0.475\textwidth}
				\centering
				\includegraphics[width=\textwidth]{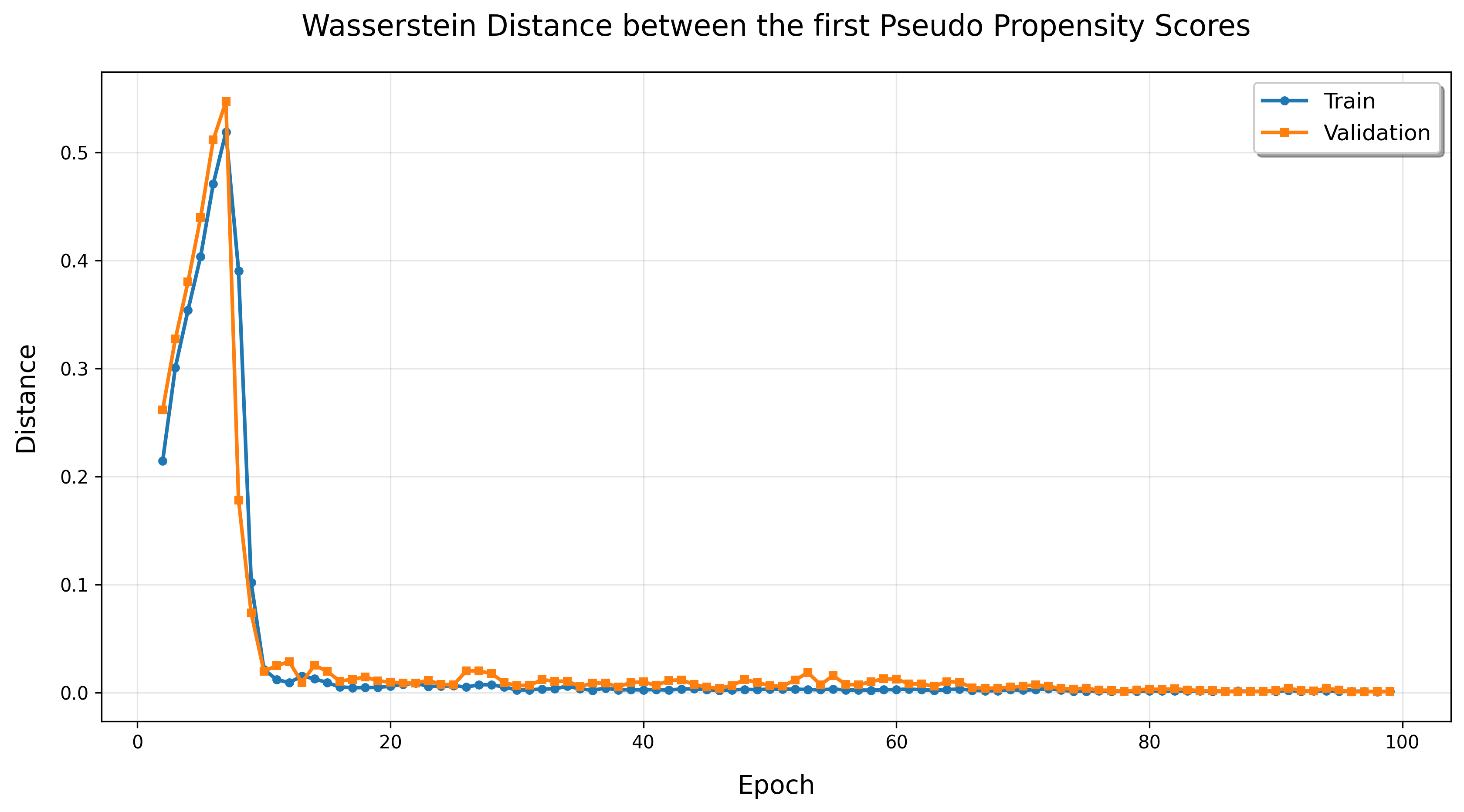}
				\caption{For the first pseudo propensity score}
				\label{case3snnnetdpp:subfig1}
			\end{subfigure}
			\hfill  
			\begin{subfigure}[b]{0.475\textwidth}
				\centering
				\includegraphics[width=\textwidth]{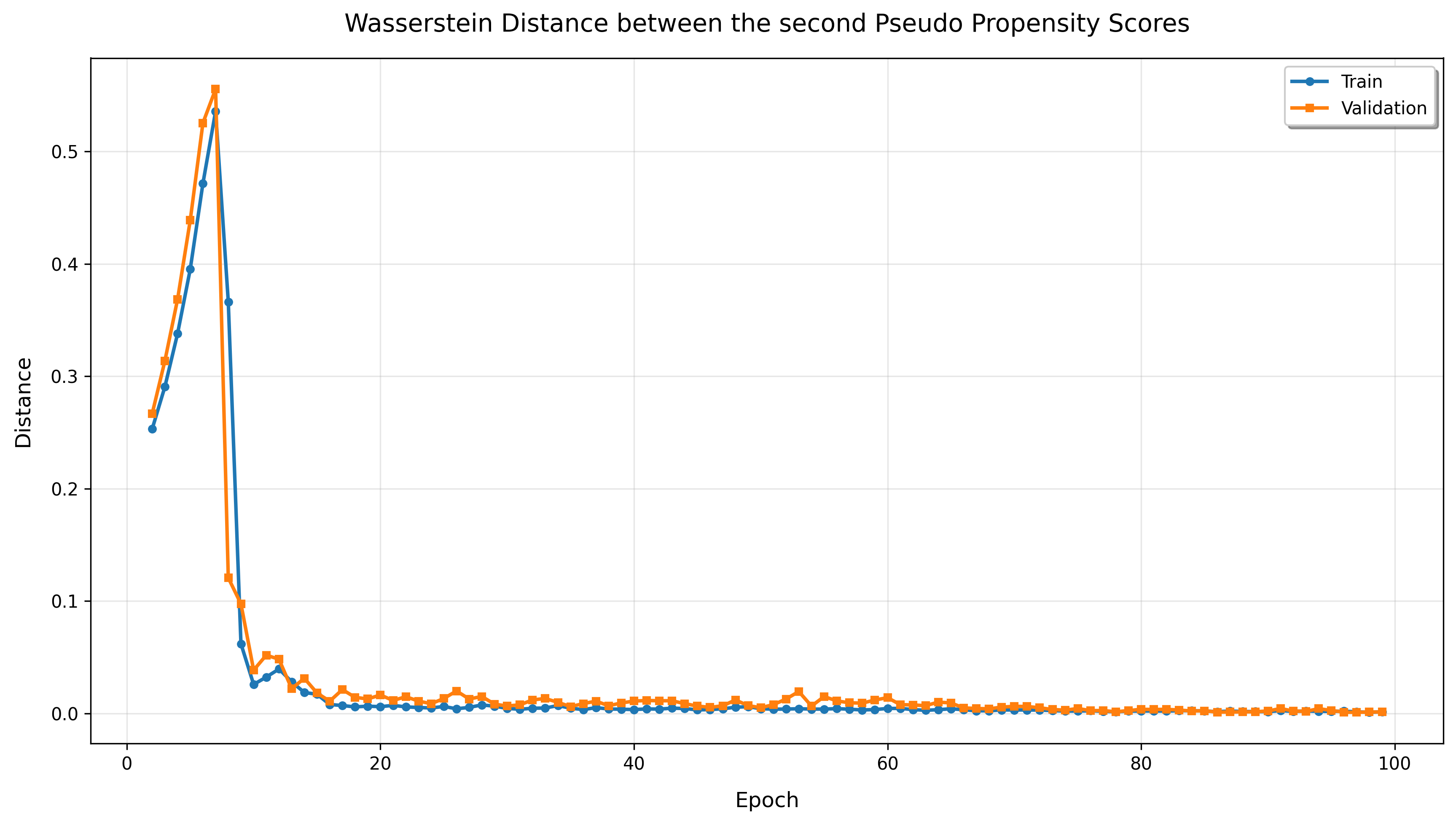}
				\caption{For the second pseudo propensity score}
				\label{case3snnnetdpp:subfig2}
			\end{subfigure}
			
			\caption{Wasserstein distances pattern between constitutive pseudo propensity scores obtained from self-balancing with two balancing nets (for data generation under Case 3).}
			\label{case3snnnetdpp:mainfig}
		\end{figure*}
		
		The distance between constitutive pseudo propensity scores are computed and plotted  to demonstrate how they behave during training periods. In Figure \ref{case3sbnnet:mainfig}, one can see the distance results and how the loss changes over time for the Sbnet architecture during a specific data generation process in Case 3.  The plots indicate that the changes in distance measurements begin to stabilize during the final epochs of training. During these epochs, the reduction in distance measures occurs simultaneously with the reduction in loss. This indicates that when the model is trained well, the convergence of the pseudo propensity score leads to the reduction in loss.
		
		Similarly for Sbnet with two pseudo propensity scores, Figure \ref{case3snnnetdpp:mainfig} shows how the distance measures change over epochs. When the models begin to train effectively, both distances of the pseudo propensity scores start converging to smaller possible values. The graphs show stability around the tails of the plots. This evidence indicates that when double propensity scores are used, the Sbnet will also be able to find stable pseudo propensity scores by using balancing nets, which helps it to balance its outcome nets.

		In general, these simulation studies show that the proposed self-balancing neural network is robust to the strength of confounding variables and the complexity of the treatment model. In the study, two aspects of self-balancing neural networks are presented: single balancing net (single pseudo propensity score is used) and double balancing nets (pseudo propensity scores are used). Convergence of the balancing nets to obtain a certain optimal pseudo propensity score is presented by showing how the Wasserstein distance behaves between consecutive pseudo propensity scores over training periods. In the following section, the application of Sbnet is shown on real-world problems.

		\section{Application}
		\label{sec5}
		In this section, self-balancing neural networks are used to estimate the average treatment effect for a real data problem. Similarly, in the simulation study, all other existing methods are also used to estimate the average treatment effect, and the results are compared. Based on graphical illustration of how Wasserstein distances between consecutive pseudo propensity scores behave over training epochs, convergence of pseudo propensity scores of the proposed Sbnets is shown.
		
		\subsection{Description of the data sets}
		To show applications of self-balancing neural networks and compare them with other existing methods on real-world problems, two datasets are used. The first dataset is the Infant Health and Development Program (IHDP). It is semi-simulated data whose outcomes are synthetic and covariates are real. This data is mainly proposed to compare causal inference models as a benchmark. It was initially used in \citet{hill2011bayesian}. The data set involves 25 health indicators, binary treatment, and continuous covariates\footnote{The dataset are publicly available. The synthetic outcomes were regenerated multiple times. The dataset used in this study can be downloaded from \href{https://github.com/AMLab-Amsterdam/CEVAE/blob/master/datasets/IHDP/csv/ihdp_npci_1.csv}{IHDP Dataset} }. 
		
		The second data set is about the effect of college education on earnings. The data set is available in the \textsf{causaldata} package for R, Python, and Stata. The dataset includes variables such as log wages (which are used to measure earnings), years of education, work experience, race (indicating whether the person is Black), residence in the southern United States, marital status, urban or rural classification based on Standard Metropolitan Statistical Areas, and whether a four-year college exists in the county. To create a binary treatment variable, years of education are classified as either greater than 12 or not. Individuals with more than 12 years of education are categorized as having attended college. 
		
		For college data, variables such as residence can affect both college education, which is the treatment variable, and wages. Such variables, therefore, become confounding variables. 
		
		In the following subsection, using these two data sets, the average treatment effect is estimated using Sbnet and other existing methods. The results obtained using the proposed method are compared with the results obtained with the existing method.
		\subsection{Results}
		
		For both data sets, a similar Sbnet architecture in the simulation setting is used, except that for the IHDP data set, instead of a mean squared error, a mean absolute error is found to be an appropriate loss. Using these architectures, the average treatment effects are estimated for both data sets.
		
		First, the average treatment effect of the Infant Health and Development Program (IHDP) set is estimated. For this data set, the true average treatment effect can be known in advance, as the outcome of the data set was synthetic. Hence, the true ATE is computed and found to be equal to 4.01. Table \ref{dataanalysis} presents the results of the estimated average treatment effect as well as the corresponding standard deviation for this data, as calculated by both proposed and existing methods. It reveals that Sbnet yields satisfactory results. Existing methods also show satisfactory results; only Kernel DML estimation highly deviates from the true ATE.

		\begin{table*}[h]
			\centering
			\small  
			\setlength{\tabcolsep}{1mm}  
			\caption{Average treatment effect estimates for application data}
			\begin{tabular}{lrrrr}
				\toprule
				\multirow{2}{*}{\textbf{Method}} & \multicolumn{2}{c}{\textbf{IHDP Data}} & \multicolumn{2}{c}{\textbf{College Data}} \\ \cmidrule(lr){2-3} \cmidrule(lr){4-5}
				&Training ATE&Test ATE &Training ATE&Test ATE\\
				\midrule

				DoWhy PMS       & 4.041 (0.045)&        4.049 (0.133)    &  0.169 (0.208) &         0.174 (0.182)\\
				DoWhy IPW      &  4.069 (0.028) & 4.074 (0.116)     &0.138 (0.007) &         0.135 (0.029)\\
				KernelDML       &  3.629 (0.103)&         3.631 (0.104) &0.207 (0.012)&        0.207 (0.012)\\
				CausalForestDML &  3.853 (0.073) & 3.853 (0.086)&     0.261 (0.024) &      0.262 (0.028)\\
				DragonNet     &3.884 (0.049) & 3.878 (0.094)       &0.198 (0.044) &  0.198 (0.045)\\
				TARNet        &  3.879 (0.055) &  3.872 (0.094)&      0.185 (0.036) &  0.185 (0.036)\\
				XGBTRegressor   &3.993 (0.015) &  4.007 (0.052) &     0.185 (0.013) & 0.182 (0.017)\\
				LRSRegressor   &  3.942 (0.023) &       3.942 (0.023)&    0.210 (0.007) & 0.210 (0.007)\\
				MLPTRegressor   &   4.518  (0.130) & 4.513 (0.166)&    0.214  (0.035) & 0.214 (0.038) \\
				\textbf{Sbnet }   &4.010 (0.039)&  4.005 (0.057)     &   0.226 (0.012) &   0.226 (0.013)\\
				\bottomrule
			\end{tabular}
			\begin{tablenotes}
				\item \footnotesize Note: The value in the table displays the estimated average treatment effect (ATE) along with the standard deviation obtained through bootstrapping. The estimated standard deviations are presented in the brackets.
			\end{tablenotes}
			\label{dataanalysis}
		\end{table*}
		
		For the second real data set, the average treatment effect of education levels on log wages is estimated. The estimated results and corresponding estimated standard deviations are presented in Table \ref{dataanalysis}. For this dataset, since the true average treatment effect is unknown, one can compare the consistency of the ATE estimated by the proposed Sbnet with estimated results obtained by existing methods. Furthermore, one may rely on the training and test results of the estimated ATE to detect whether the model is overfitted or not.

		\begin{figure}[h]
			\centering
			
			\begin{subfigure}[b]{0.475\textwidth}
				\centering
				\includegraphics[width=\textwidth]{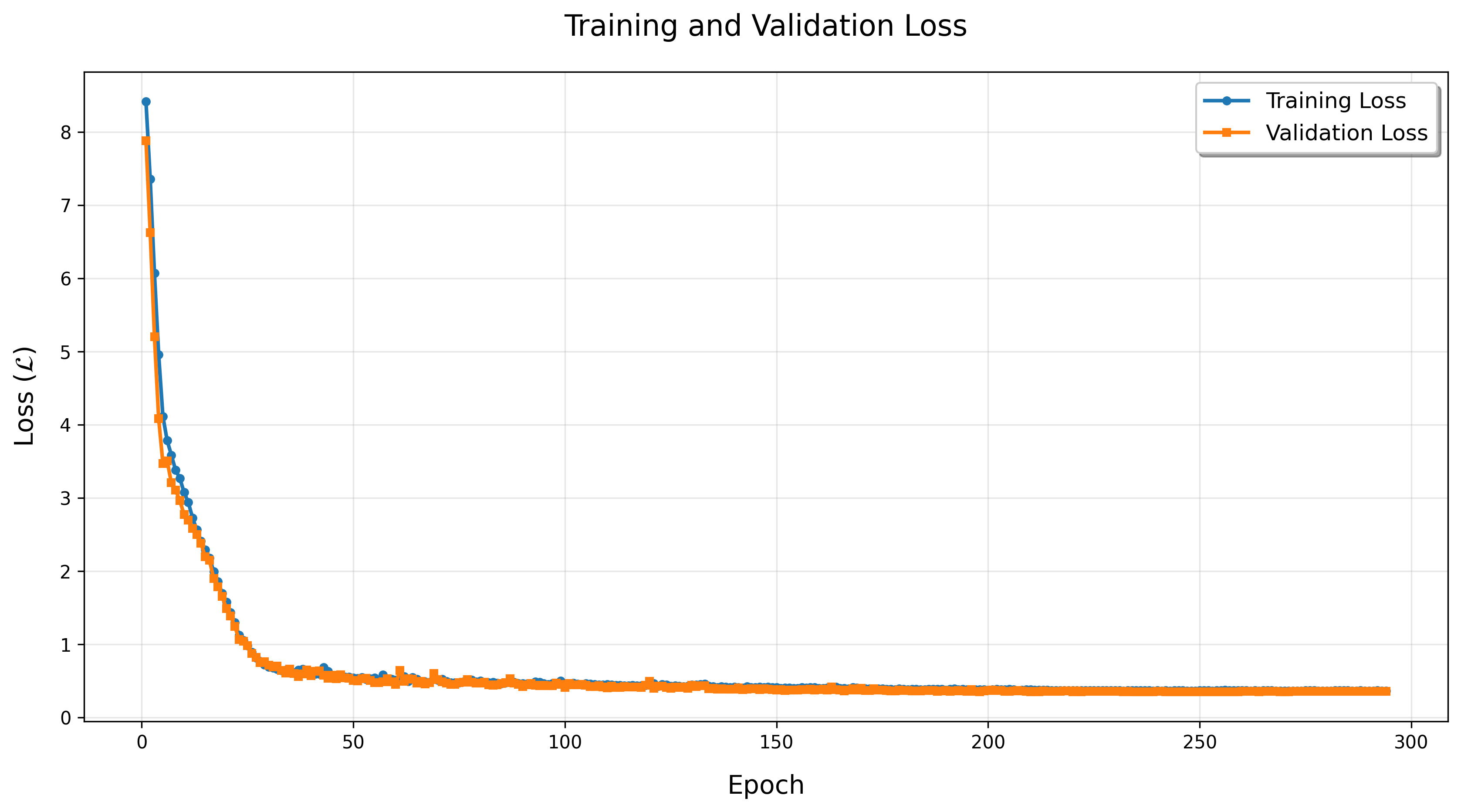}
				\caption{Loss}
				\label{ihdp:subfig1}
			\end{subfigure}
			\hfill  
			\begin{subfigure}[b]{0.475\textwidth}
				\centering
				\includegraphics[width=\textwidth]{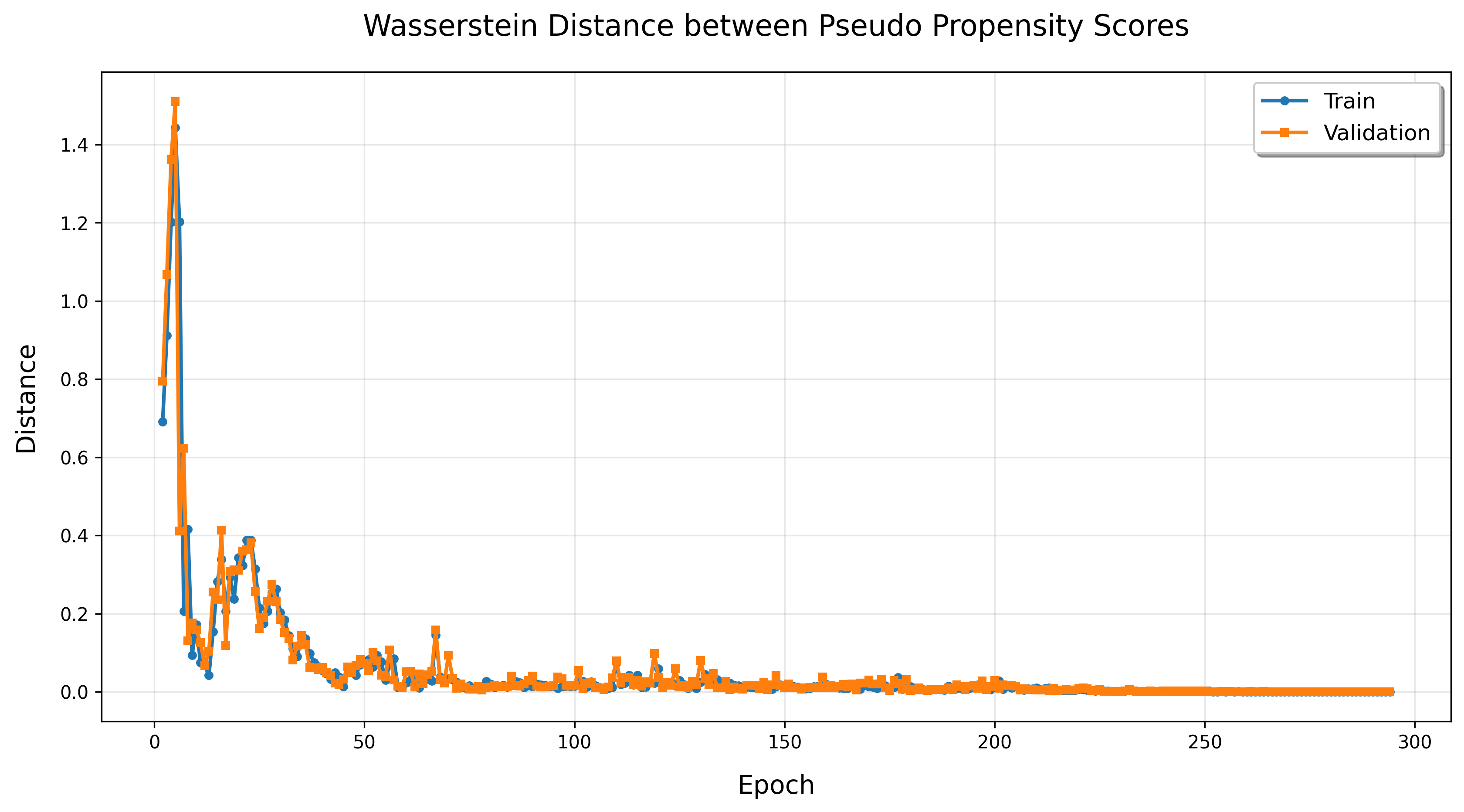}
				\caption{Wasserstein Distance}
				\label{ihdp:subfig2}
			\end{subfigure}
			
			\caption{Pattern of loss in estimating outcome and Wasserstein distance between consecutive propensity scores (for IHDP dataset). The loss is computed for all iterations while the computation of Wasserstein distance starts at the second iteration.}
			\label{ihdp:mainfig}
		\end{figure}
		
		The estimated average treatment effect by the proposed self-balancing neural network for both training and test data was 0.226. This indicates that the model does not have an overfitting or underfitting problem. Additionally, the estimated standard deviations of the estimated ATE for both training and test data are 0.012 and 0.013, respectively, indicating that they are nearly equal. Moreover, one can see that the estimated ATE by Sbnet is in agreement with the result obtained from existing methods.
		
		Finally,  the convergence of the pseudo-propensity score is graphically represented by the Wasserstein Distance measure between consecutive evaluations. For the IHDP dataset, when Sbnet with a single pseudo propensity score is used, as shown in Figure~\ref{ihdp:mainfig}(\subref{ihdp:subfig2}), the result of the Wasserstein distance initially shows irregular behavior, then as loss in Figure~\ref{ihdp:mainfig}(\subref{ihdp:subfig1}) starts to converge, the Wasserstein distance measure converges. This indicates that the consecutive pseudo propensity scores are becoming increasingly stable during training and ultimately converging. 
		
		In general, the proposed method for estimating ATE has shown satisfactory performance on real-world problems. It is in agreement with the results obtained in the simulation study. Hence, the proposed one-step method proposed to estimate ATE is a reasonable approach to handle both balancing and estimation of average treatment effect simultaneously.
		\section{Conclusion}
		
		\label{sec6}
		
		In summary, this paper presents a novel method to estimate the average treatment effect of a self-balancing neural network architecture. This design helps the neural network by generating its propensity score, which is called the pseudo propensity score. Unlike other methods that depend on propensity scores, the proposed method doesn't need a separate model to obtain a propensity score to balance covariates or for regression adjustment. Rather, it is an integrated part of the model. 
		
		This approach enables the estimation of the average treatment effect using a single neural network model that incorporates structural adjustment. Such formulations help to do further analysis, such as ensemble learning, which was part of our proposed work to introduce the multi-pseudo propensity score, which is an entirely new concept, and analyzing the importance of pre-treatment effect and others. 
		
		The current paper focuses on estimating the average treatment effect of binary treatment on continuous outcome. The proposed method shows promising results in handling average treatment effect estimation. One can enhance the proposed model to estimate treatment effects when the outcome is binary, multinomial, or involves multiple treatments. Moreover, one can improve the model to handle the heterogeneous treatment effect problem.

		\section*{Acknowledgements}
This study is funded by the Shanghai Natural Science Foundation (24ZR1420400), National Natural Science Foundation of China (12401347 and 72331005), the Basic Research Project of Shanghai Science and Technology Commission (22JC1400800), and the Shanghai "Science and Technology Innovation Action Plan" Computational Biology Key Program (23JS1400500 and 23JS1400800).

		\section*{Disclosure statement}
		There is no author(s) reported potential conflict of interest.
		
		
		\bibliographystyle{elsarticle-harv} 
		\bibliography{sbnetreferences}
		

	\end{document}